\documentclass{article}
\usepackage{fancyhdr}

\usepackage[numbers,sort&compress]{natbib}
\usepackage[final]{neurips_2026}
\usepackage{newtxtext}
\newcommand{\WorldStopProb}{{\omega}}
\newcommand{\MaxWorldStopStep}{{M}}
\newcommand{\SelfStopStep}{{S}}
\newcommand{\WorldStopAtIProb}{\widetilde{\omega}} 
\newcommand{\DontKnowProb}{d}
\newcommand{\SelfStopProb}{s}
\newcommand{\discount}{\gamma}
\newcommand{\NextTokenProb}{t}
\newcommand{\DesiredStopProb}{\rho}
\usepackage{tikz}
\usetikzlibrary{positioning, arrows.meta, calc} 

\usepackage[utf8]{inputenc} 
\usepackage[T1]{fontenc}    
\usepackage{hyperref}       
\usepackage{url}            
\usepackage{booktabs}       
\usepackage{amsfonts}       
\usepackage{nicefrac}       
\usepackage{microtype}      
\usepackage{xcolor}         
\usepackage{graphicx}
\usepackage{enumitem}
\usepackage{amsmath}
\usepackage{subcaption}
\usepackage{xspace}
\usepackage{letterspace}
\usepackage{multirow}
\usepackage{array}
\definecolor{darkblue}{HTML}{000090}
\usepackage{ragged2e} 
\usepackage[export]{adjustbox} 
\usepackage{makecell}

\hypersetup{
    colorlinks=true,
    citecolor=darkblue,
    linkcolor=darkblue,
    filecolor=darkblue,      
    urlcolor=darkblue,
}

\newcommand{\comment}[3]{\textcolor{#1}{[\textbf{\textsc{#2}} #3]}}
\def\@noticestring{}

\usepackage[textsize=tiny]{todonotes}
\newif\ifdraft
\drafttrue
\ifdraft

    \newcommand{\marginvn}[1]{\todo[color=red!20,size=\tiny]{ VN:  #1}}
    
        \newcommand{\mcm}[1]{\comment{blue}{MCM:}{#1}}
\newcommand{\alex}[1]{\comment{red}{Alex}{#1}}

\else
    \newcommand\marginvn[1]{}
    \newcommand{\inlinevn}[1]{}
\newcommand{\mcm}[1]{}
\newcommand{\alex}[1]{}
    \newcommand\note[1]{}
    \presetkeys{todonotes}{disable}{}
\fi

\usepackage[capitalize,noabbrev]{cleveref}

\title{Catch Your Breath: Adaptive Computation for Self-Paced Sequence Production}

\title{Catch Your Breath: Adaptive Computation\\ for Self-Paced Sequence Production}
\author{Alexandre Galashov\thanks{Address correspondence to \texttt{mcmozer@google.com}}\\
  Google DeepMind
  \And
  Matt Jones\\
  Google DeepMind
  \And
  Rosemary Ke\\
  Google DeepMind
  \AND
  Yuan Cao\\
  Google DeepMind
  \And
  Vaishnavh Nagarajan\\
  Google DeepMind
  \And
  Michael C.~Mozer\\
  Google DeepMind 
}

\begin{document}

\maketitle
\newcommand{\pause}[1][]{{\normalsize\textls[-20]{\textsc{<pause#1>}}}\xspace}
\newcommand{\dk}{{\normalsize\textls[-20]{\textsc{<don't know>}}}\xspace}
\newcommand{\ldp}{{\ensuremath{\ell_\text{CYB-DP}}}\xspace}
\newcommand{\lrho}{{\ensuremath{\ell_\text{CYB}\DesiredStopProb}}\xspace}
\newcommand{\lva}{{\ensuremath{\ell_\text{CYB-VA}}}\xspace}
\newcolumntype{M}[1]{>{\centering\arraybackslash}m{#1}}

\begin{abstract}
Within the landscape of inference-time scaling methods for foundation models, a \emph{width-based} approach to scaling---which involves the insertion of \pause tokens in the input stream to delay model responses---offers a unique advantage by increasing model expressivity while remaining highly parallelizable at both training and inference. The existing literature on training models to utilize \pause tokens relies on the standard cross-entropy objective in which the model output is read out and evaluated only at the final step of a pause sequence. This approach provides no mechanism for the model to regulate its own processing or to signal readiness to respond, treating the additional compute steps as a barrier to skip over rather than a resource to be used adaptively. We propose a supervised loss, \emph{Catch Your Breath} (\emph{CYB}), framed as a sequential-decision problem, that trains a model to dynamically and autonomously scale the number of compute steps used for each input token. The model indicates the need for additional compute steps by emitting a special \dk output, delaying its response via a pause. The model can abstain multiple times to obtain longer delays. Our experiments demonstrate that when \pause tokens are included in the input stream, pretraining or fine-tuning with CYB significantly outperforms standard cross-entropy,
reducing perplexity and enhancing downstream accuracy with no additional computational or memory cost.

\end{abstract}

While remarkable advances in AI have been achieved by scaling model size and dataset size,
a third scaling dimension has recently become a focus of intense interest:  
scaling the computational budget during inference.
The most straightforward approach to inference-time scaling involves `thinking' in some form,
whether chain of thought \citep[e.g.,][]{zelikman2022,wei2023,yao2023}, repeated
sampling, self critiques, or latent-thought methods
\citep[e.g.,][]{phan2023,hao2024,ruan2025}. 

Inference-time scaling can also operate at a more granular level by
boosting resources devoted to individual tokens,  either by expanding model \emph{depth} or 
model \emph{width}.  Depth-based approaches involve layer recurrence, i.e., repeating 
individual layers or a range of layers \citep[e.g.,][]{giannou2023}.
Width-based approaches involve inserting special \pause tokens into the input stream \citep{goyal2024}.
Each \pause informs the model that it has an additional forward pass to perform further
processing on the context and potentially improve the next token emitted. 

Although depth-based approaches have gotten the bulk of notice in the literature (see 
Section~\ref{sec:related_work}), our work focuses on width-based approaches which offer 
a complementary form of expressivity \citep{london2025,merrill2025} and can be parallelized,
unlike depth-based approaches.
However, width-based approaches come with a critical drawback which our work aims to address.
In width-based approaches,  processing time is expanded for
an input token by inserting one or more trailing \pause tokens, and when these tokens occur,
the model output is ignored until the final pause step \citep{goyal2024}. To appreciate 
the computational challenge this insertion poses for a transformer, consider an indexed token 
sequence such as 
\vspace{-.12in}
{\begin{center}{\texttt{0:Donald, 1:lost, 2:his, 3:<pause>, 4:<pause>, 5:mind}.}\end{center}}
\vspace{-.12in}
Had there been no pauses here, the model would have been trained at step 2 to predict \texttt{mind}; with the pauses at steps 3 and 4, the model is trained to output \texttt{mind} at step 4 instead. However, this causes two troubles at inference time. 
First, even if the model already predicts well at step 2, it must hold its response for two additional steps. Additional computation time is not always helpful.
Second, because the model does not control the number of pauses, it may act greedily and try to predict at each step, which may deny it the opportunity to plan optimally with the additional pauses in cases where more compute is beneficial.

\begin{figure}[b!]
    \centering
    \includegraphics[width=\linewidth]{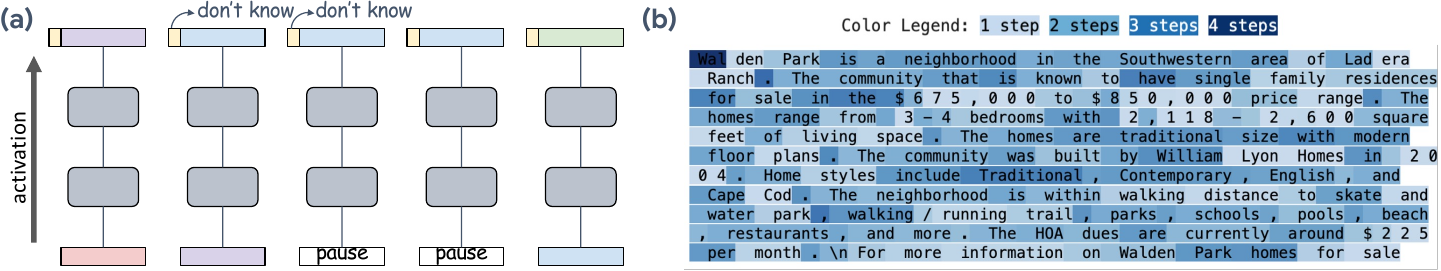}
    \caption{
    (a) Sequence processing in a transformer with \dk outputs that trigger \pause inputs.
    (b) The CYB loss yields variable per-token computing steps.
    When the \dk probability is high, additional computing steps are provided.
    Shading indicates the expected number of computing steps to predict the corresponding token.
    Spacing reflects token boundaries.
    \label{fig:fig1}}
\end{figure}

To remedy these limitations of externally imposed delays, we propose a method 
enabling the model to autonomously insert variable-duration delays as needed.
This method is inspired
by studies of human reading, which find that readers fixate longer on words when processing demands rise 
(see Appendix~\ref{sec:howpeopleread}). 
For each input token, the model is given the option
of emitting a special \dk token to abstain from responding. With abstention,
the arrival of further input is held off and a \pause token in the stream signals
continued processing (Figure~\ref{fig:fig1}a).
The model can produce abstention responses multiple times in a row, 
yielding adaptive self-paced sequence production (Figure~\ref{fig:fig1}b).

To summarize our contributions: 
(1) We propose a principled loss, called \emph{Catch Your Breath} or \emph{CYB}, 
that trains a transformer to self-determine how many computation steps it requires for
each input token.  While most adaptive-computation techniques focus on inference 
acceleration, we show that training pause tokens with CYB yields uniformly superior 
performance compared to the existing (non-adaptive) method in the literature---applying the standard cross-entropy objective on the last pause step.
(2) We propose a parallel training procedure for 
the CYB loss that is as compute and memory efficient as 
cross entropy.
(3) We explore variants of the loss that include additional penalties for slow responses 
but find---consistent with our conjecture---that the CYB loss naturally drives the model to respond efficiently. 
(4) We show significant advantages of the CYB loss over cross entropy, both for 
models pretrained from scratch and models fine-tuned with \pause tokens, on 
both perplexity measures and 
downstream model performance. 
(5) We investigate where the model chooses to pause, and find
systematic and intuitive insights into the model's behavior.

\section{Related research}
\label{sec:related_work}
\emph{Pause tokens.}
Prior work has proposed special content-free input tokens to improve model performance, variously
referred to as `memory' \citep{burtsev2021}, `thinking' \citep{herel2023}, `dummy' \citep[][Appendix A.3]{lanchantin2023}, `registers' \citep{darcet2024}, and `filler' \citep{pfau2024} tokens.
\citeauthor{goyal2024} \citep{goyal2024} coined the term `pause' tokens and focused on their role in granting models additional computation steps. They were the first to conduct extensive experiments with pretraining and fine-tuning, and subsequent work has utilized their paradigm. We will refer to their
approach by the title of their paper, \emph{Think Before You Speak} or \emph{TBYS}.
During pretraining, their data loader inserts $\MaxWorldStopStep$ pauses into the text stream 
at uniformly random positions.  When pauses are inserted, model outputs for the 
$\MaxWorldStopStep-1$ steps leading up to the final pause token are ignored and 
excluded from the usual next-token-prediction cross-entropy loss. During fine-tuning, 
where a prefix sequence is given along with a target sequence, $\MaxWorldStopStep'$ pauses are 
appended to the prefix, giving the model additional compute steps before its response. 
\citeauthor{goyal2024} found consistent benefits in downstream-task accuracy only when pauses
were included in both pretraining and fine-tuning stages.  Although this approach appears to 
be an alternative to chain-of-thought methods, pauses may be related given that substituting 
meaningless filler tokens for thinking still yields benefits \citep{pfau2024}.
In previous work, long delays are introduced between a prefix and a model's response, intended to 
support explicit reasoning about and calculation of the response---an alternative to chain of thought. 
In contrast, our motivation is to allow short delays in processing any token, with the aim 
of facilitating integration of information in prefix tokens as well as in choosing any token in the response.
Similar in spirit to our proposal,  Dong et al. \citep{dong2025reinforcementpretraining} propose an RL method that
generates reasoning traces for individual tokens, also aimed at next-token prediction.
The computational cost of our technique is significantly lower.

\emph{Anytime prediction and adaptive computation.}
In anytime prediction, models make a series of responses that progressively improve
with additional compute. Such models exhibit a \emph{speed-accuracy} trade-off in which a coarse answer appears quickly
but is refined with additional computation. In some schemes, the model is rewarded for getting the right response in
as few steps as possible \citep[e.g.,][]{iuzzolino2021}, similar to adaptive computation time (ACT) for 
recurrent networks \citep{graves2016}.  However, whereas ACT and anytime prediction focus 
on dynamically terminating inference to reduce wall-clock latency, our approach uses a sequential-decision framework 
to adapt computation within a fixed parallel-step budget.
PonderNet \citep{banino2021} applies probabilistic halting over serial depth recurrence
and minimizes expected loss regularized by a geometric KL prior, whereas CYB operates over
parallel input steps and minimizes log of expected discounted accuracy without a KL penalty
(though see Appendix~\ref{app_sec:DPVA}).

\emph{Adaptive depth models.} 
Extending adaptive computation-time to transformers, \emph{looped transformers} and variants have 
been proposed in which the functional depth of a model can be varied by repeating individual layers or ranges of layers
\citep{yang2024,nowak2024,alabdulmohsin2025recursive,bae2025,chen2025,geiping2025,koishekenov2025,li2025skiplayerloopit,mcleish2025,rodkin2025,yu2025backattn,zhu2025,zeng2026ponderlm,jeddi2026loopformer,ng2026rys}.
Adaptation can be made on a per-token basis with a learned router.
Looping increases model depth, whereas pause tokens increase model width (input steps). Because depth increases require 
serial computation, looping cannot exploit hardware parallelism like pause tokens can. Looping and pausing extend model expressivity 
in complementary ways \citep{london2025,merrill2025}, and thus should be considered independent possibilities rather than 
competing alternatives.

\emph{\dk outputs.}
The \dk output is related to classification with abstention (or rejection) \citep{bartlett2008,liu2019,hendrickx2024}. 
Two recent articles introduce heuristic notions of ``I don't know'' to determine where pause tokens should be inserted.
\citet{cohen2024} include a \dk output and introduce a 
heuristic training objective that shifts probability to the \dk output when the
model predicts incorrectly. Their approach applies multiple heuristics to prevent
the degenerate solution of always outputting \dk. By integrating the \dk option into
our sequential-decision task, our approach does not suffer such degenerate solutions.
\citet{kim2025} propose a heuristic method, DIT, for dynamically inserting pause tokens based on 
low token likelihood (confidence).
Their training procedure relies on this proxy as the basis for inserting pauses, whereas CYB 
learns where additional compute actually improves accuracy via a principled sequential-decision loss.
DIT is based on a two-pass training procedure; in contrast, CYB requires only a single pass, albeit with 
a larger number of \pause tokens.
Finally, DIT cannot distinguish aleatoric from epistemic uncertainty, 
which the CYB loss can in principle infer.
Because DIT and CYB differ on so many dimensions, a compute-and-memory-matched comparison is highly challenging.

\section{Optimizing compute steps with the Catch-Your-Breath (CYB) loss}
\label{sec:main_method}

We propose a loss that lets a transformer \emph{autonomously} determine the number of compute steps used for individual tokens. (If notation in this section is challenging, Table~\ref{table:notation} provides a glossary.)

In prior work with pause tokens, each input token is allotted up to $\MaxWorldStopStep$ compute steps 
by inserting between 0 and $\MaxWorldStopStep-1$ \pause tokens into the input stream. Specifically, an external
controller, which we refer to as the \emph{world}, draws the allotted number of steps 
$i \in  \{1, 2,  \ldots, \MaxWorldStopStep \}$  from a prior distribution $\boldsymbol{\omega} = (\omega_1, \ldots,
\omega_\MaxWorldStopStep )$, where $\omega_i$ is the probability that a token is granted exactly $i$
steps ($i-1$ pauses) to respond. A training loss is applied at the final step.
Under causal masking, however, the model cannot anticipate how many pauses follow the current step. It must
therefore be prepared to respond at each step, 
leading to greedy optimization that may not best leverage the allocated compute time.
Following \citep{goyal2024}, we refer to this state-of-the-art baseline approach as 
\emph{Think Before You Speak} (\emph{TBYS}).

We hypothesize that pause steps might be better utilized if the model itself can control whether pauses are used. 
To this end, our approach, CYB, lets the model autonomously expand the processing of each token up to as many as 
$\MaxWorldStopStep$ steps.
However, the model can stop sooner: at each step $i<\MaxWorldStopStep$,
the model can choose either to respond or to abstain via a \dk response.
If the model chooses to respond or $i=\MaxWorldStopStep$ (the deadline is reached),
a token is drawn from the usual next-token distribution, excluding \dk.

\begin{figure}[tb!]
\centering
\includegraphics[width=.90\linewidth]{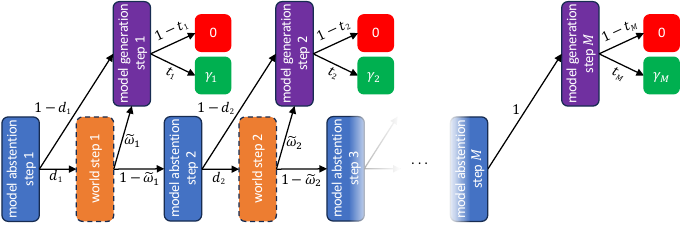}
\caption{
Catch Your Breath: A sequential-decision framework to obtain a single output token. The blue boxes represent 
model abstention decisions, the purple boxes model choices of output tokens. The orange boxes are choices made by the 
environment about whether the model is granted a delay in responding. The green and red boxes represent end-state
rewards. 
The unconditional stop-probability at step $i$, $\omega_i$, has been rewritten as a hazard probability 
$\tilde{\omega}_i = {\WorldStopProb_i}/{\textstyle \sum_{j=i}^{\MaxWorldStopStep} \WorldStopProb_j}$.
\label{fig:mdp}}
\end{figure}

We frame the selection of each generated token as a sequential-decision problem, 
depicted in Figure~\ref{fig:mdp}. 
The subscripts in the Figure are a step index relative to the presentation of the real input token, $i \in \{ 1, 2, \ldots, \MaxWorldStopStep \}$. The model has a sequence of decision points, 
indicated by the blue and purple boxes. At each step $i$, the model selects \dk with probability $\DontKnowProb_i \in [0,1]$, a value that is dependent on model parameters and input. In accordance with the
model's confidence in its answer (probability $1-\DontKnowProb_i$), it selects an output token. The
selected token will be either the target, with probability $\NextTokenProb_i$, or an
incorrect response, with probability $1-\NextTokenProb_i$. The green squares indicate the reward (accuracy) when correct, discounted
by $\discount_i \in [0,1]$ at step $i$, and the red squares are the non-reward when incorrect. 

Our sequential-decision framework allows for the external world to terminate processing early
(as in TBYS). It incorporates components,
depicted by the dashed orange boxes in Figure~\ref{fig:mdp}, which permit the world
to terminate processing under the prior distribution $\boldsymbol{\omega}$.
In most experiments we conduct with CYB, models are allowed to use $\MaxWorldStopStep$ steps per 
token, i.e., $\boldsymbol{\omega} = (0,0,\ldots,1)$.

For a given token $k$, where the model produces \dk outputs $\boldsymbol{\DontKnowProb}^k = 
(\DontKnowProb_1^k, \ldots, \DontKnowProb_M^k)$, the step at which the model generates an output---self- or 
world-induced---is a random variable, $\SelfStopStep$, with probability given by
\vspace{-.02in}
\begin{equation} \textstyle
\Pr(\SelfStopStep=i \mid \boldsymbol{\DontKnowProb}^k,\boldsymbol{\WorldStopProb}) =
\left( \WorldStopProb_i + (1-\DontKnowProb_i^k)\sum_{j=i+1}^{\MaxWorldStopStep} \WorldStopProb_j \right) \prod_{j=1}^{i-1} \DontKnowProb_j^k
\label{eq:stoptime}
\end{equation}
and  $\DontKnowProb_{\MaxWorldStopStep}^k \equiv 0$ for the final step. 
Equation~\eqref{eq:stoptime} follows from Figure~\ref{fig:mdp}; Appendix~\ref{app_sec:derivation_output_prob}
has a derivation.

For token $k$, with model stepwise prediction accuracy $\boldsymbol{\NextTokenProb}^k = 
(\NextTokenProb_1^k, \ldots, \NextTokenProb_M^k)$, the expected discounted accuracy in the CYB decision task is $\mathbb{E}_{i \sim  \Pr(\SelfStopStep \mid \boldsymbol{\DontKnowProb}^k,\boldsymbol{\WorldStopProb})}
\left[ \discount_i \NextTokenProb_i^k \right]$,  
which can be used to define a generalized negative log likelihood (NLL) loss:
\vspace{-.02in}
\begin{equation}
\ell_\text{CYB}(\boldsymbol{t}^k, \boldsymbol{d}^k;~ \boldsymbol{\omega}, \boldsymbol{\gamma}) = -\log 
\mathbb{E}_{i \sim 
\Pr(\SelfStopStep \mid \boldsymbol{\DontKnowProb}^k,\boldsymbol{\WorldStopProb})
} \left[ \discount_i \NextTokenProb_i^k \right] .
\label{eq:cyb}
\end{equation}
Averaging over all tokens $k$, we obtain the standard expectation-of-log NLL formulation: 
\begin{equation}
\mathcal{L}_\text{CYB} = \mathbb{E}_k \left[
\ell_\text{CYB}(\boldsymbol{t}^k, \boldsymbol{d}^k;~ \boldsymbol{\omega}, \boldsymbol{\gamma})\right] .
\label{eq:loss}
\end{equation}
This loss is used to optimize model parameters via back propagation through the outputs $\boldsymbol{t}^k$ and the \dk probabilities $\boldsymbol{d}^k$. It is conditioned on hyperparameter vectors
$\boldsymbol{\WorldStopProb}$, the distribution over the maximum number of steps allotted for 
a token, and $\boldsymbol{\discount}$, the step-wise discounting vector.

TBYS \citep{goyal2024}, the standard cross-entropy loss for \pause tokens, can be formulated as
\begin{equation}
\label{eq:tbys_loss}
\ell_\text{TBYS} (\boldsymbol{t}^k; \boldsymbol{\omega})  = \mathbb{E}_{i\sim \text{Cat}(\boldsymbol{\omega})} \left[
-\log t_i^k
\right],
\end{equation}
where the stop-probability priors, $\boldsymbol{\omega}$, correspond to the manner in which 
tokens are  inserted into the stream for training. For example, if two pauses are randomly 
inserted following 10\% of real tokens, $\MaxWorldStopStep=3$ and $\boldsymbol{\omega} = [.9,0,.1]$.
In our implementation, TBYS is trained with the stop-prior distribution provided to CYB,
which simplifies the loss to 
$\ell_\text{TBYS} \left(\boldsymbol{t}^k \mid \boldsymbol{\omega}=(0,0,\ldots,1)\right)  =  -\log t_M^k$.

CYB is motivated by a set of hypotheses, \textbf{H1}-\textbf{H4}, that we evaluate via simulation experiments.
\begin{description}[leftmargin=!,labelwidth=\widthof{\bfseries H3.}] 

\item[H1.] We conjecture that CYB will outperform TBYS when matched in training data and compute resources
because CYB self-determines when to read out whereas TBYS lacks this
control.
\item[H2.] CYB should yield a well calibrated model: to minimize the loss, the \dk probability at step
$i$, $d_i$, should be low when the target probability, $t_i$, is high and vice-versa.
Further, the model should be time efficient: to minimize the CYB loss, the model should respond at 
step $i$ unless the model's response is likely to improve at step $i+1$. Thus, it should not 
request all $\MaxWorldStopStep$ steps if not warranted by improved accuracy.
\item[H3.] Our hypothesis is that CYB's benefit stems from its control over when to respond,
because having this control allows more effective use of computation. Thus,
if deadline uncertainty is added via the stop-time distribution, $\boldsymbol{\omega}$, CYB cedes some
control (like TBYS), and we predict that accuracy will drop.
\item[H4.] The model's speed-accuracy trade-off can be shifted to allow the CYB-trained model to 
respond in fewer average steps at the expense of accuracy. We can impose a linear cost per step with,
say,  $\discount_i = \discount_0^{i-1}$  for hyperparameter $\discount_0$. 
\end{description}

\vspace{-.10in}
\section{Methodology}
\vspace{-.10in}

\textbf{Position encoding with pause sequences.}
To encode an input sequence, each token is assigned a position index for use by positional embedding methods 
such as RoPE \citep{su2023}. Treating \pause inputs like any other token will cause them to shift the position 
indices of following tokens. This approach was used by \citet{goyal2024}, but may be problematic when introducing 
pauses during fine-tuning in a model that was trained without them. 
Consequently, we use RoPE but assign \pause tokens the same index as the immediately preceding non-\pause token. 
To ensure that the model can selectively index a specific pause token, we encode each pause in the sequence with a unique
token code: \pause[1], \pause[2], etc. 

\textbf{Representing abstention.}
To obtain the \dk output, $\DontKnowProb_i$, we repurpose an unused token $\delta$ from the standard set of output
alternatives and then renormalize the model's output distribution, $y$, as $\hat{y}_{i,j} = y_{i,j} / (1-y_{i,\delta})$ 
for token indices $j \ne \delta$, and $\hat{y}_{i,\delta} = 0 $ otherwise. Whether fine-tuning or pretraining, we
wanted to ensure a large enough \dk prior probability that the model would learn to use it. We renormalized model posteriors
to reflect a high prior for \dk, as detailed in Appendix~\ref{app_sec:methods}.

\textbf{Efficient training with CYB.} Although the sequential-decision framework (Figure~\ref{fig:mdp}) might suggest training
the model over sequential steps, we propose a procedure in which $\MaxWorldStopStep-1$ pauses are inserted into
the input stream after each real input token during training.  The transformer steps are run in parallel with 
causal masking. The pause steps may or may not be used by CYB, e.g., if at some step $i$ the model 
is certain ($\DontKnowProb_i \to 0$) or the world enforces a stop (via $\WorldStopProb_i$), the outputs at subsequent pauses 
will not contribute to the current token's loss.
During inference, when the transformer might be run autoregressively, we match the use of pauses at train time, 
i.e., always inserting $\MaxWorldStopStep-1$ pauses after each real input token. Because an input token and the pauses that 
follow can be processed in parallel, there is a trivial computational cost to inserting the maximum number of pauses in the stream for every
token. Rather, the potential downside comes from the pauses cluttering the context window. 





%

\vspace{-.09in}
\section{Results}
\vspace{-.10in}
\label{sec:results}

We conducted experiments using Gemma1-2B~\citep{gemmateam2024gemmaopenmodelsbased} and Gemma3-4B~\citep{gemmateam2025gemma3technicalreport}, which we fine-tuned on a subset of C4 
\citep{c4_dataset}. We use sequences of 2048 real tokens which we expand to context windows of 4096, 6144, and 8192 tokens by inserting
1, 2, or 3 pauses following each real token, respectively.
Unused token indices were repurposed to serve as pauses and \dk. Unless explicitly mentioned, all experiments conducted below use fine-tuned Gemma1-2B.
Additionally, we conducted a pretraining experiment with Gemma1-2B using C4
\citep{c4_dataset}.
Further details of the training procedure can be found in Appendix~\ref{app_sec:methods} and extended results in
Appendix~\ref{sec_app:additional_results}.

\begin{figure}[t!]
     \centering
     \begin{subfigure}[t]{0.48\textwidth}
         \centering
         \includegraphics[width=\linewidth]{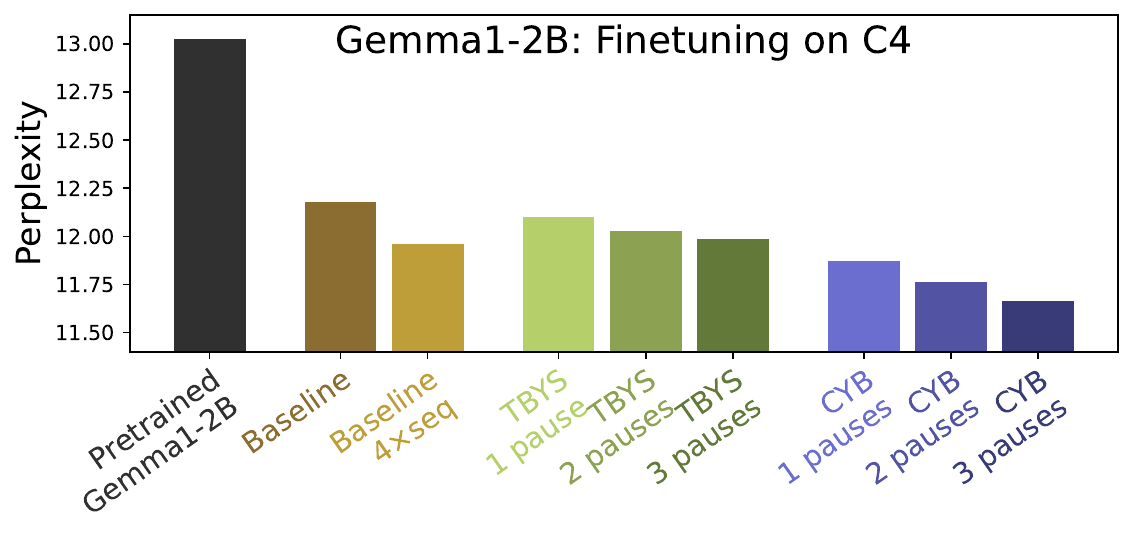}
     \end{subfigure}
     \hfill 
     \begin{subfigure}[t]{0.48\textwidth}
         \centering
         \includegraphics[width=\linewidth]{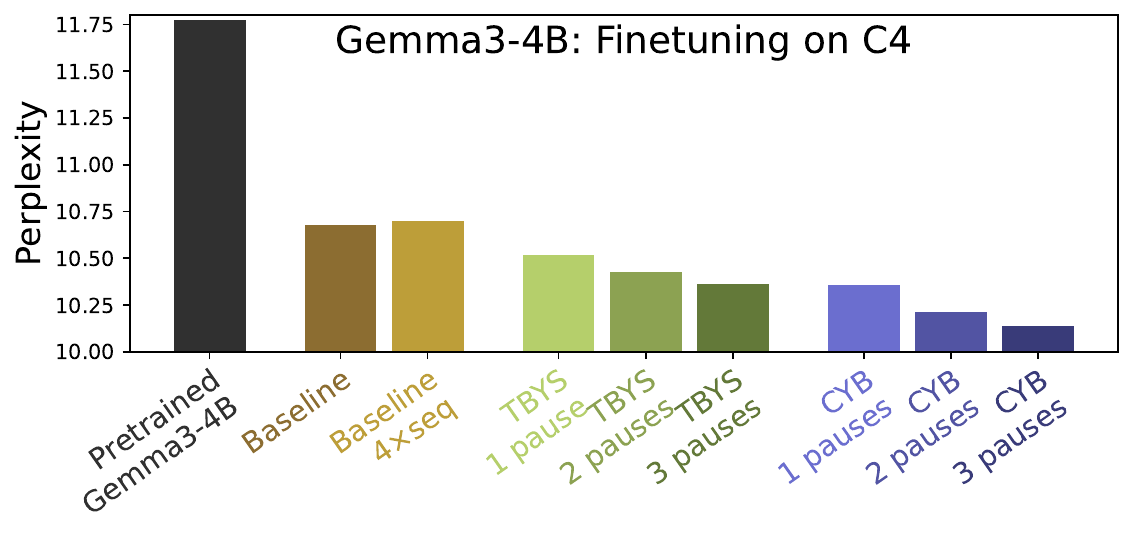}
     \end{subfigure}
     \caption{C4 evaluation set perplexity (lower is better) for pretrained Gemma1-2B (\textbf{left}) and Gemma3-4B (\textbf{right}) models (black bar) and various models fine-tuned on the C4 train set. 
Model variants include a baseline condition, where no modification is made to 
training objective or token sequence (brown bar), Think Before You Speak (TBYS) 
with 1, 2, and 3 pause tokens after each input token (green bars), and the CYB with 1, 2, and 3 pause tokens.}
     \label{fig:finetuning_results}
\end{figure}




Our experiments focus on comparing two alternative losses for training a model with input \pause tokens: CYB, our method, and TBYS \citep{goyal2024}, which uses the standard cross-entropy loss and reads out the model's response at the final \pause. We match CYB and TBYS in the insertion
of pauses, with $n \in \{1,2,3\}$ pauses after each real token. The methods differ only in the loss and the token-readout procedure. 
Compute and memory requirements are identical.
Additionally, we train baseline methods on the same data \emph{without the pauses} in order to show that pausing
improves model quality. We can either match the baseline and pausing methods (CYB, TBYS) in terms of training data or compute.
To match training data, we train a model referred to as \emph{baseline} with a 2048 token context window.
To match compute of the three-pause CYB and TBYS models, we train a model referred to as \emph{baseline 4$\times$seq} with an 8192 token context window which 
is filled with real  data, providing it with fourfold advantage in data over the pausing methods.


\paragraph{Fine-tuning experiments.} Figure~\ref{fig:finetuning_results} presents perplexity on a C4 evaluation set 
across different fine-tuned model variants. As one would expect, the pretrained model with no fine-tuning (black bar) has higher perplexity than 
the baseline model fine-tuned on C4 (brown bar). The light brown bar is a second baseline trained on sequences of length 8192 real tokens, which is 
compute matched to TBYS and CYB with three pauses.  CYB consistently outperforms TBYS, to the point where 1-pause CYB beats 3-pause TBYS.
The data-matched baseline (dark brown bar) has higher perplexity than any of the CYB and TBYS methods; the baseline that is compute matched to 
3-pause TBYS/CYB (light brown bar) has higher perplexity than CYB with 1, 2, or 3 pauses.

To address reliability concerns, we fine-tuned a baseline Gemma3-4B model to the C4 dataset.
We performed three replications of three conditions: baseline (no pauses), CYB with 3 pauses, and TBYS with 3 pauses. 
We found almost no variance in eval performance across seeds for baseline
(10.670, 10.670, 10.669), CYB (10.131, 10.134, 10.129), or TBYS (10.359, 10.360, 10.358).
These replications vary only the training data order since the weights come from the pretrained model.

Curiously, \citet{goyal2024} found that TBYS did not improve beyond baseline with fine-tuning.
The difference might be explained by several adjustments to their methodology.
We assign pause tokens distinct
identities depending on their relative positions, and we use the same absolute sequence
position for all pauses in a row. Further, we insert pause tokens systematically
after every true input token, whereas \citeauthor{goyal2024} inserted them
randomly and infrequently.


\begin{figure}[b!]
     \centering
     \begin{minipage}[t]{0.48\textwidth}
         \centering
         \includegraphics[width=\linewidth]{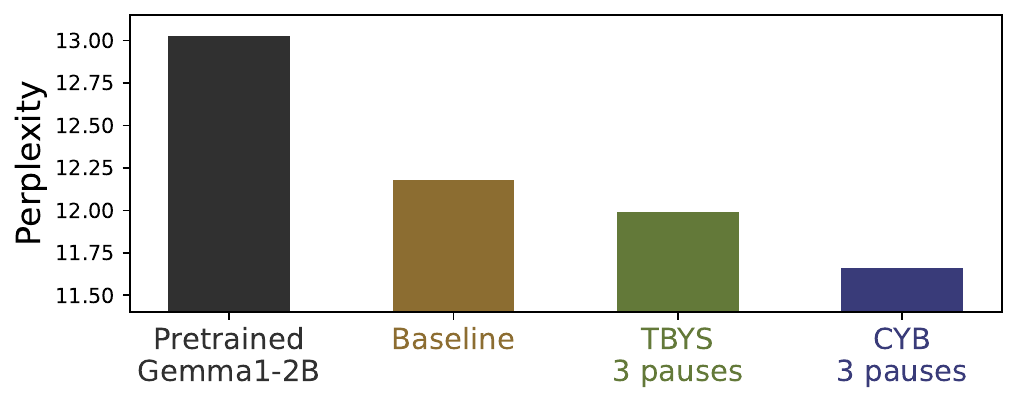}
         \caption{C4 evaluation set perplexity for Gemma1-2B architectures pretrained from scratch on C4 (except for the black bar, which uses open-source weights). Model variants include a baseline condition without pauses and models with three pauses per real token:
         TBYS \citep{goyal2024} and CYB (ours).}
         \label{fig:pretraining_performance}
     \end{minipage}
     \hfill 
     \begin{minipage}[t]{0.48\textwidth}
         \centering
         \includegraphics[width=\linewidth]{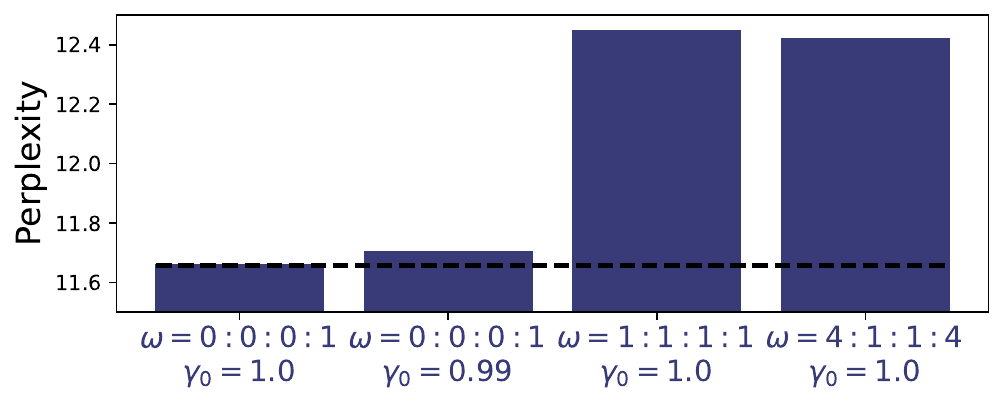}
         \caption{C4 evaluation set perplexity for various hyperparameters of CYB with Gemma1-2B fine-tuned on C4. Hyperparameter $\omega$ refers to the distribution of stopping times and is indicated by ratios for steps 1-4 (i.e., the initial input token and the following 3 pause tokens). Hyperparameter $\gamma_0$ refers to an exponential discount factor. }
         \label{fig:cyb_hparams}
     \end{minipage}
\end{figure}

\paragraph{Pretraining experiment.} We also conducted a pretraining experiment on C4, comparing TBYS and CYB with 3 pauses, with baselines included
as a reference  (\cref{fig:pretraining_performance}). As with fine-tuning experiments, we observe superior performance by both TBYS
and CYB over the baselines, and CYB delivers the strongest performance. CYB shows a larger benefit over TBYS in the pretraining setting relative
to the fine-tuning setting: 3.6\% lower perplexity for pretraining versus 2.7\% for fine-tuning. (See Tables~\ref{tab:gemma1_2b_finetuning} and \ref{tab:gemma1_2b_pretraining} in Appendix~\ref{sec_app:tabular_representation}.)
Both the fine-tuning and pretraining experiments support our hypothesis \textbf{H1} that CYB outperforms TBYS. These methods are matched in memory, data, and computation and differ only in the loss function and the token-readout procedure.



\begin{figure}[t!]
    \centering
    \begin{minipage}[c]{0.6\linewidth}
        \centering
        \includegraphics[width=\linewidth]{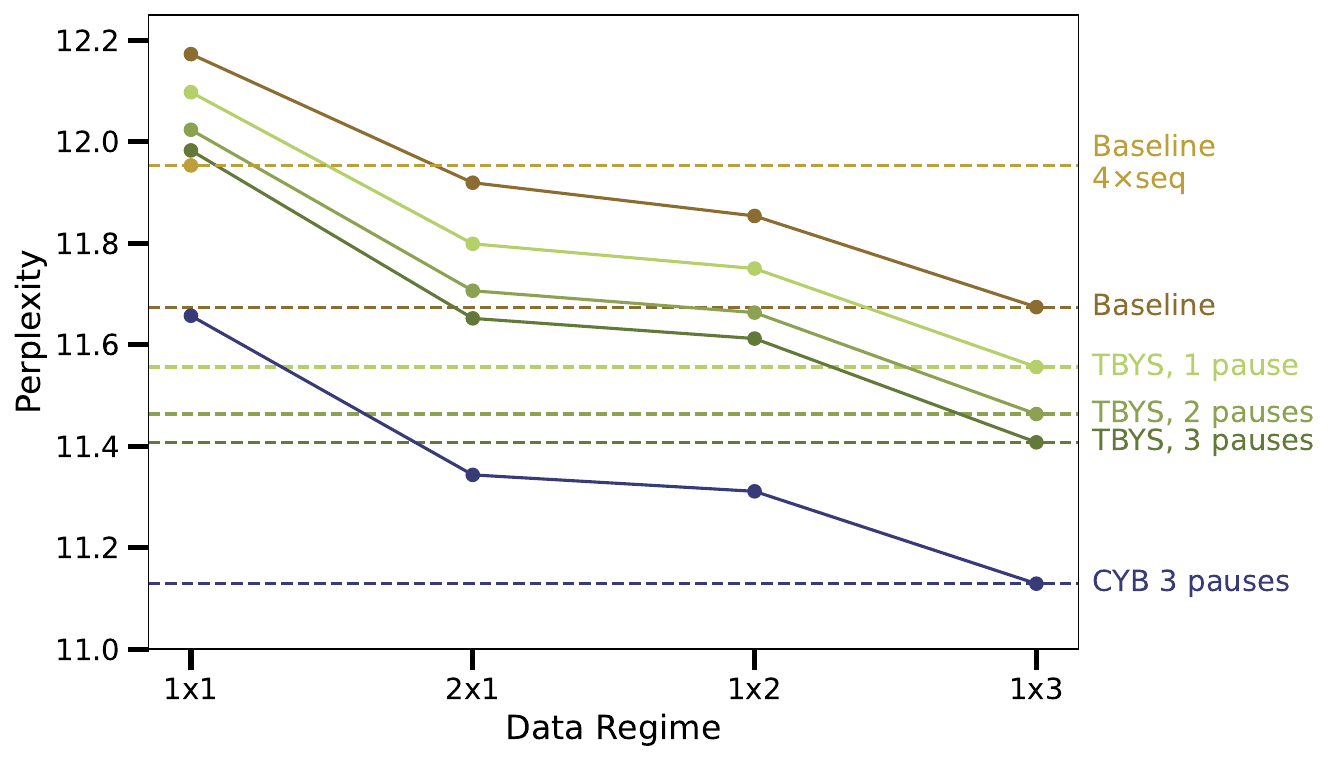} 
    \end{minipage}
    \begin{minipage}[c]{0.37\linewidth}
        \caption{C4 evaluation set perplexity for various fine-tuned Gemma1-2B models based on a given data regime. 1$\times$1 and 2$\times$1 refer to 1 and 2 passes through the fine-tuning set, respectively. 1$\times$2 and 1$\times$3 refer to a single pass through 2 or 3 times the dataset size of 1$\times$1.  The dashed lines indicate the perplexity of each model in the 1$\times$3 training regime, except for \emph{Baseline 4$\times$seq} where we only report performance in the 1$\times$1 regime.
        \label{fig:data_regimes}}
    \end{minipage}\hfill
\end{figure}

\paragraph{Data efficiency of CYB.} While CYB reduces perplexity, developing intuitions
about the significance of this reduction is challenging. We aim to offer such intuition by 
casting the improvement in terms of data efficiency. Figure~\ref{fig:data_regimes} presents evaluation perplexity 
of various models fine-tuned in different data regimes. The regimes labeled {1$\times$1} and 
{2$\times$1} correspond to one and two passes through our 6.4M training sequences, respectively. The regimes 
labeled {1$\times$2} and {1$\times$3} correspond to data sets that are two and three times as large, 
respectively.   CYB requires just 33\% of the data to match baseline performance, and less than
66\% of the data to beat performance of TBYS with three pauses. Interestingly, the performance gap between CYB 
and the alternative methods remains intact as data sets grow, indicating CYB's gains are a genuine improvement 
in learning efficiency, not an artifact of data saturation.
This experiment finds further support for hypothesis \textbf{H1}, the superiority of CYB over TBYS, here in data efficiency.

\vspace{-.10in}
\paragraph{Uniform-stopping ablation.}
CYB differs from TBYS in two respects: it incorporates learnable abstention and it trains the output token on each pause step.
We have hypothesized that learned abstention is critical (\textbf{H3}). To support this hypothesis, we conducted an ablation in which 
learned abstention is eliminated. We imposed a uniform distribution over stopping step via the hyperparameter 
vector $\boldsymbol{\omega} = (0.25, 0.25, 0.25, 0.25)$ and trained a variant of 
CYB with a maximum of $M=4$ steps, ignoring the \dk output. 
This ablation that trains the correct response at each pause step performed extremely poorly: on Gemma3-4B, C4 evaluation 
perplexity is 18.60 for the ablation versus 10.13 for CYB. The ablated model did improve reliably on the training set over fine 
tuning, but showed the classic sign of overfitting on 
the evaluation set: perplexity dropped and then ramped back up; the lowest perplexity achieved was 13.67. Thus, we find evidence that 
allowing the model to control the processing flow with \dk outputs drives performance gains, distinguishing CYB from 
adaptive-computation time or anytime-prediction methods.

\vspace{-.10in}
\paragraph{Hyperparameter exploration.} We explored two hyperparameters of CYB introduced in 
Section~\ref{sec:main_method}: $\boldsymbol{\omega}$ and $\gamma_0$. 
Instead of allowing the three-pause CYB model to use as many steps
as it wished, we imposed an external stop time distribution. We use the notation
$\boldsymbol{\omega} = \omega_1:\omega_2:\omega_3:\omega_4$ to denote the ratio of probabilities of
forcing the model to stop after processing steps 1-4. When CYB has full determination of
when to stop within the pause sequence,  $\boldsymbol{\omega}=$ 0:0:0:1.
The other hyperparameter, $\gamma_0$, controls the exponential discounting of accuracy at each step
(i.e., accuracy is discounted by $\gamma_0^{i-1}$ at step $i$); $\gamma_0=1$ for no discounting.

Turning first to $\boldsymbol{\omega}$, Figure~\ref{fig:cyb_hparams} shows perplexity for fine-tuned 
models with various hyperparameters and 3 pauses per token. The leftmost bar imposes no early stopping and no 
discounting---the result we previously presented. The rightmost two bars indicate performance when early stopping
is imposed, either uniformly ($\boldsymbol{\omega}=$ 1:1:1:1) or with a bimodal distribution ($\boldsymbol{\omega}=$
4:1:1:4). Consistent with \textbf{H3}, which posits that CYB's advantage over TBYS stems from having control over
when it responds, we see that losing this control via early stopping severely harms its performance. 

\begin{figure}[t!]
    \centering
    \includegraphics[width=.9\linewidth]{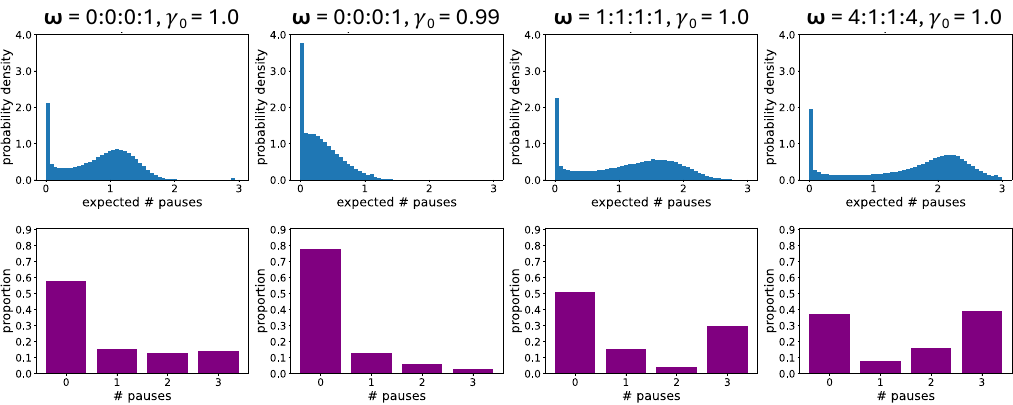}
    \caption{{Latency distribution for four hyperparameter settings of CYB}. Blue histograms show the distribution over evaluation tokens of the expected latency (the number of pause steps utilized). Purple bar graphs show the latency
    distribution averaged across tokens.}
    \label{fig:steps_ap}
\end{figure}

In contrast, when exponential discounting ($\gamma_0=0.99$) is applied, only a small cost in perplexity arises, as represented
by the second bar from the left in Figure~\ref{fig:cyb_hparams}. Nonetheless, discounting is very effective in
reducing the expected stopping step of the model, as indicated by the distribution shift between the first two columns 
of Figure~\ref{fig:steps_ap}. To explain these distributions, CYB's \dk output allows us to compute the
\emph{latency} distribution (Equation~\ref{eq:stoptime}), i.e., the distribution over which step the model
reads out its response.  Figure~\ref{fig:steps_ap} presents statistics of the latency distribution for the four 
hyperparameter settings of CYB previously described in Figure~\ref{fig:cyb_hparams}. The blue curves are a 
probability density over the expected latency for individual tokens. The purple bars show the latency 
distribution averaged over tokens. The distributions in the first column ($\gamma_0=1$) shift leftward in
the second column ($\gamma_0=.99$), consistent with hypothesis \textbf{H4} that we can manipulate the model's 
speed-accuracy trade-off with discounting. With discounting, delaying a response is penalized unless the delay results
in a corresponding boost in accuracy.

In the leftmost column of Figure~\ref{fig:steps_ap}, the model has no discounting penalty ($\gamma_0=1$) and 
need not accommodate forced stops ($\boldsymbol{\omega}=$ 0:0:0:1), yet
CYB is nonetheless judicious in its use of pauses.
This finding seems surprising given that additional pause steps should not 
\emph{harm} performance. However, the finding is consistent with hypothesis \textbf{H3}, 
which asserts the advantage of CYB over TBYS is that CYB can stop when it is ready to respond
instead of holding its response until the final step. 

Paradoxically, forcing the model to stop early shifts its preferred latency distribution slower,
as seen in the last two columns of Figure~\ref{fig:steps_ap}. Perhaps the increased likelihood 
of \dk is simply a reflection of its worse performance (Figure~\ref{fig:cyb_hparams}).
We explored using soft constraints (secondary losses) to encourage the model to adhere to a 
target stop-time distribution, as an alternative to discounting. However, this approach
yielded strictly worse performance from CYB. See Appendix~\ref{app_sec:DPVA} for details.

\paragraph{Calibration of model confidence and accuracy.}  
If successful, CYB should train models to self-calibrate (Hypothesis \textbf{H2}): \dk should 
have high probability when a delay in responding will improve model output. 
Appendix~\ref{app_sec:calibration_theo} provides a theoretical argument for this conjecture
by showing that model gradients depend on improving accuracy over steps. 
Appendix~\ref{app_sec:calibration_emp} presents empirical evidence that 
the model implicitly predicts its own accuracy at subsequent steps and is sensitive to 
this prediction in using the \dk output. A statistically reliable positive correlation 
is found between \dk probability and an accuracy gain with additional processing for all steps 
of a three-pause model, supporting the calibration hypothesis \textbf{H2}.

\paragraph{Downstream task performance.} We evaluate Gemma3-4B models fine-tuned on C4 with three pauses per token (zero for
the baseline) on several downstream benchmarks: HellaSwag 10-shot~\citep{zellers2019hellaswag}, MMLU 5-shot~\citep{hendrycks2021mmlu}, and GSM8K 8-shot Chain of Thought (CoT) ~\citep{cobbe2021gsm8k}, reporting top-one accuracy in all cases. In these experiments, we insert three pauses after every real token for CYB and TBYS, including
tokens in the prompt and in the response, at both fine-tuning and evaluation. For CYB, we generate responses by deterministic
selection of the most probable response token, including \dk. This procedure means that we run the sequential-decision framework
selecting the most probable choice at every stage.
For GSM8K, the C4 fine-tuned models achieved $0\%$ accuracy in a zero-shot setting; we therefore additionally fine-tuned each 
model on the GSM8K training set for 2 epochs with a learning rate of $10^{-4}/\!\sqrt{d_{\text{model}}}$ and a warmup 
with $18$ steps. We conducted ten replications with different random seeds which determined batch composition and ordering. Results are summarized in Table~\ref{tab:downstream} and match results of the perplexity analysis: CYB yields a consistent improvement in accuracy  over TBYS, and both beat the baseline no-pause model. 
For the comparison of TBYS and CYB on GSM8K, we observe a reliable improvement in accuracy with CYB over TBYS (two-tailed
paired t-test: $t(9)=2.85$, $p=0.019$).

\begin{table}[t]
  \centering
  \caption{Downstream evaluation top-1 accuracy (\%).
    HellaSwag and MMLU are evaluated on C4 fine-tuned checkpoints.
    GSM8K results are means and standard errors over 10 replications in which we additionally fine tune on the GSM8K training set.}
  \label{tab:downstream}
  \small
  \begin{tabular}{lccl}
    \toprule
    \textbf{Method} & \textbf{HellaSwag} & \textbf{MMLU} & \textbf{GSM8K} \\
                    & (10-shot)          & (5-shot)      & (8-shot)       \\
    \midrule
    Baseline                    & 24.68 & 49.16 & 47.60 \\
    TBYS                        & 24.74 & 52.70 & 53.16 \tiny[SEM 0.23] \\
    CYB               & \textbf{27.94} & \textbf{54.00} & \textbf{53.91} \tiny[SEM 0.30] \\
    \bottomrule
  \end{tabular}
\end{table}

\begin{table}[t!]
    \centering
    \caption{Statistics of pause counts for selected tokens following CYB training}
    \label{table:pause_stats}
    {\small
    \begin{tabular}{|c|>{\Centering\arraybackslash}m{2.25in}|>{\Centering\arraybackslash}m{2.25in}|} 

    \hline
    & \textbf{Low Median} & \textbf{High Median} \\
    \hline
    \makecell{\textbf{Low}\\\textbf{Variance}} & 
    isn, wasn, didn, doesn, don, according, able, However, plenty, etc, Inc, \&, non, addition
    & 
    devices, players, students, projects, applications,
    challenges, environment, systems, stories, games,
    groups, events, patients, families, teams
    \\
    \hline
    \makecell{\textbf{High}\\\textbf{Variance}} & 
    g, e, to, entry, S, won, co, New, Y, V, F, Z, World, of, Air
    & 
    materials, images, @, site, pictures,
    photos, sites, code, College, wood,
    page, file, professionals, is, homes
    \\
    \hline
    \end{tabular}
    } 
\end{table}

\paragraph{Qualitative results.} Table~\ref{table:pause_stats} shows pause-count statistics for selected 
tokens for the three-pause CYB model. The Table presents tokens which tend to be followed  by few 
(\emph{low median}) or many (\emph{high median}) pauses, and those which tend to have the same 
number of pauses regardless of their context (\emph{low variance}) or  have context-sensitive pause 
durations (\emph{high variance}). There are clear regularities in these tokens, such as the upper
right cell being plural nouns, the upper left cell having highly predictable completions, and the lower left cell
containing single letters.

Figure~\ref{fig:fig1}b
presents a sample evaluation token sequence. Each token's background is colored to indicate the expected number of pause steps requested by the model in order to predict that token. (That is, the coloring reflects the processing steps conditioned on the preceding token.) The coloring ranges from white
to dark blue, corresponding to zero pause steps (1 total step) to
3 pause steps (4 total steps).
The most obvious feature of the pause pattern is that it is not uniform
but varies on a token-by-token basis. For example, the pause following
the token `traditional' in Figure~\ref{fig:fig1}b tends to be long; in contrast,
the model does not slow down when it is predicting a string of digits.
More examples of shaded text can be found in Supplementary Figures~\ref{fig:test_example2} and \ref{fig:test_example3}.

\section{Discussion}

We introduced Catch Your Breath (CYB), a sequential-decision framework and loss function that supports
inference-time scaling of per-token computation.  CYB has the option of utilizing additional compute steps when the output 
is uncertain, allowing it to pause and delay its output. In classical classification 
with abstention, epistemic uncertainty stems from data scarcity or gaps in training coverage. Under the CYB 
objective, however, uncertainty is fundamentally a reflection of computational constraints: the model is 
uncertain not because it lacks representation in its weights, but because it has not yet performed
sufficient computations to integrate long-range context.  As our calibration metrics show, 
subsequent pause steps allow the model to resolve this internal uncertainty using test-time compute.

We described CYB as performing \emph{width-based} scaling, which is an underexplored approach relative 
to \emph{depth-based} scaling (e.g., looped transformers).
While both width-based and depth-based scaling of computation can be performed without an increase in the memory footprint
of the model weights, width-based scaling has a much lighter impact on inference latency. The two forms of scaling 
can be combined, though, and extend model expressivity in complementary ways \citep{merrill2025}.

The existing paradigm used for width-based scaling, which we refer to as Think Before You Speak (TBYS), is 
a limiting case of CYB suffering from the fact that it lacks autonomy over the final readout point. We
presented strong evidence that CYB outperforms TBYS (hypothesis \textbf{H1}), and that CYB's benefit stems from
its control over when to respond (hypothesis \textbf{H3}). Our focus on comparing CYB and TBYS allows for a strict,
rigorously matched comparison: the two methods have identical memory and computation requirements at both training and inference;
they differ only in the loss function used and the readout point. The benefits for CYB over TBYS are essentially cost free.

Although we demonstrated that CYB can be used to control the speed-accuracy trade-off (hypothesis \textbf{H4}),
it is important to emphasize that CYB is not motivated by raw inference-time compute savings; rather, it is about 
providing the architecture with flexible, context-dependent control over its computation.
Just as human readers selectively modulate fixation durations to resolve lexical ambiguities or integrate 
complex discourse (see Appendix A), CYB-trained models naturally learn to pause only when the local transition dynamics 
demand it. This behavior is conceptually aligned with looped transformers that utilize internal routing components to decide whether or not to continue depth recurrence. However, width-based adaptation through CYB offers distinct structural and engineering advantages over these depth-recurrent models, including simplicity and elegance of the routing rule, parallelizability, and formal expressivity.

\textbf{Limitations and future directions.} The recipe we used for variable computation in CYB involves inserting the maximum
number of pauses after each real token. While this recipe is highly parallelizable and introduces no computational or 
memory overhead \emph{relative to TBYS}, it can limit the effective sequence length for long-form generation. 
One might
imagine that running the model stochastically---inserting a \pause only when an explicit
\dk is triggered---would be interesting, but the only significant benefit of stochastic
rollouts would be in KV cache memory, given that multiple pauses can be inserted and computed
in parallel. 
Recent work suggests an alternative approach to reducing the memory footprint.
In a particular setting, \citet{monea2026} and \citet{langford2026} found that 
benefits of pausing can be obtained without storing all pause steps in KV cache.
This approach ought to be tested for CYB.

Other potential directions for future research include:
\begin{itemize}
    \item 
\emph{Non-causal processing of the pauses.} Since \pause tokens are placed in the input stream simultaneously, they can inform
one another as activation is propagated forward.
Then the CYB loss trains the model which \pause token output to use for its response.
\item 
\emph{Selective insertion of pauses}. Instead of placing a sequence of pauses after every real token, one could instead selectively
insert the pauses at positions most likely to benefit from them,
as determined by analysis of trained models (e.g., Table~\ref{table:pause_stats}).

\item
\emph{Variable pause budgets.} Our work primarily inserted two or three pauses after each input token, both at training and evaluation.
There is no need for the training and generation budgets to be the same. Indeed, \citet{goyal2024} observed benefit of inserting a larger number of pauses at generation time than the model had seen during training.
\item
\emph{Looping.}
We argued that looped transformers and pausing---depth-based and width-based scaling---were independent architectural decisions 
and are motivated by different concerns. Nonetheless,
it would be valuable to study whether interactions exist; e.g., pausing may be more effective in the context of looped models.

\item 
\emph{Range of models and model scales.}
We tested only the Gemma family of models and primarily smaller models. CYB should be evaluated on other families, larger architectures, and MoE-based transformers.
\end{itemize}


\clearpage 

\begin{ack}
We are grateful to the insights and participation of Jasper Uijlings early in the course of this research, and to the anonymous reviewers and area chair who provided constructive feedback
on the research and manuscript. This research was supported by Google.
\end{ack}

%
%
\nocite{williams2026}
\bibliography{main}

\newpage

\appendix
\setcounter{table}{0} 
\setcounter{figure}{0} 

\renewcommand{\thetable}{\thesection.\arabic{table}}
\renewcommand{\thefigure}{\thesection.\arabic{figure}}

\section{How people read}
\label{sec:howpeopleread}

\begin{figure}[b!]
    \centering
    \includegraphics[width=\linewidth]{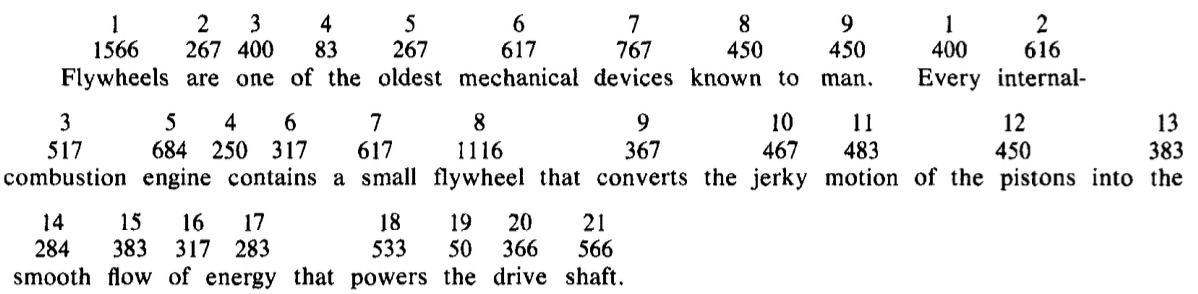}
    \caption{Gaze fixations of a human reader. Indices indicate order of gaze within
    a sentence along with fixation duration (in msec). Reprinted from \citeauthor{just1980} (\citeyear{just1980}, Figure 1).
    } \label{fig:justcarpenter}
\end{figure}

The inspiration for this research comes from studies of human reading. When gaze is
tracked as an individual reads, inter-saccade durations are highly
nonuniform. Readers fixate longer when processing demands rise, e.g., 
at points where information can be integrated across phrases, where inferences can be made 
at the end of sentences, and where low-frequency terms occur \citep{just1980}.
Figure~\ref{fig:justcarpenter} shows fixation order within a sentence for a human 
reader as well as the fixation duration. Note that short function words are skipped 
or quickly glanced over. When an uncommon interpretation of a word or sentence is 
required, readers pause. For instance, when idioms that typically have a figurative 
meaning (e.g., `break the ice') are used in a literal way, processing time slows
at the point where the disambiguating information is provided \citep{arnon2023}.
Similarly, readers pause when anomalous or unexpected information is provided,
including attributes and spatial position of a character or object \citep{stewart2009}.
These various results suggest that people integrate information in an ongoing manner,
where each incoming word and clause is processed with respect to prior context, 
and reading time is governed by the computational demands of online evaluative processes.

In humans, the ongoing, word-by-word integration processes might be described 
as \emph{microinference}, in contrast to \emph{macroinference}, which involves 
solving complex reasoning and planning problems.  In AI, past research on pause tokens
has focused on their utility for macroinference \citep{goyal2024,herel2023,pfau2024,kim2025}. 
Our present work is aimed at emphasizing the role of delays for microinference,
motivated by the need to integrate and interpret information in an ongoing fashion.
One important clue that this focus will be productive comes from a study
which found that GPT-2 attention patterns could predict human reading times,
in particular measures of the dispersion of attention and the change in attention
patterns across time steps \citep{oh2022}.

\newpage
\section{Notation used in the main article}
\begin{table}[hbt!] 
  \centering 
  \caption{Notation Used in the Main Article}
  \label{table:notation} 

  \begin{tabular}{|p{1in}|p{4.1in}|} 
    \hline 
    \textbf{Term} & \textbf{Meaning} \\
    \hline 
    $\MaxWorldStopStep$ & maximum number of steps granted by world for any token \\
    \hline 
    $\SelfStopStep$ & r.v. denoting \# steps at which model generates result (either by world or model \dk), $1 \leq S \leq  \MaxWorldStopStep$ \\
    \hline
    $\WorldStopProb_i$ & prob.\ that world terminates processing at step $i$ \\
    \hline
    $\DontKnowProb_i$ & prob.\ of model abstention (\dk) at step $i$ \\
    \hline
    $\SelfStopProb_i$ & prob.\ of outputting result (self- or world-induced) at step $i$; shorthand for $\Pr(S=i)$ \\
    \hline
    $\NextTokenProb_i$ & prob.\ that the model selects ground-truth token at step $i$ \\
    \hline
    $\discount_i$ & discount factor on accuracy at step $i$, $0 \le \discount_i \le 1$ \\
    \hline
    $\discount_0$ & hyperparameter for exponential discounting, with $\gamma_i = \gamma_0 ^ {i-1}$ \\
    \hline
    $\DesiredStopProb_i$ & [CYB-VA, CYB-DP] prior prob.\ of self-induced termination at step $i$ \\
    \hline
    $\alpha$ & [CYB-DP] penalty coefficient on prior-distribution mismatch \\
    \hline
  \end{tabular}
\end{table}

\section{Guiding CYB to obtain a desired stop-time distribution}
\label{app_sec:DPVA}
We describe two variants of CYB that impose an additional constraint on model outputs with the goal of obtaining a desired stop-time distribution. We refer to this target distribution as $\boldsymbol{\rho}$.

\subsection{Variational approach (CYB-VA)}

This approach aims to obtain a desired distribution $\boldsymbol \DesiredStopProb$ and to
minimize the loss with respect to this distribution, i.e., 
\[
\ell_{\text{CYB}\DesiredStopProb} = - \log \mathbb{E}_{i \sim \text{Cat}(\boldsymbol \DesiredStopProb)} \left[ \discount_i \NextTokenProb_i \right] .
\]
However, this loss does not depend on $\boldsymbol d$ and will thus fail to
train the \dk output, let alone train it to be consistent with $\boldsymbol \DesiredStopProb$.
To achieve this goal, we take a variational approach and train the model using the ELBO. 
Treating the model's likelihood at step $i$ as the penalized value $\Pr[{\rm target}|S=i]=\discount_i\NextTokenProb_i$, the ELBO is
the right hand side of the inequality:
\[
\begin{split}
\log \mathbb{E}_{i \sim \text{Cat}(\boldsymbol \DesiredStopProb)} \left[ \discount_i \NextTokenProb_i \right]  &\geq  
\mathbb{E}_{i \sim \text{Cat}({\boldsymbol \SelfStopProb})} \log \left[ \discount_i \NextTokenProb_i \right] - D_{\text{KL}}( \boldsymbol{s}~||~{\boldsymbol{\DesiredStopProb}}) 
\end{split}
\]
where we assume as before that externally induced stops are eliminated. 
The \dk distribution, $\boldsymbol d$, is incorporated via the stopping time distribution  $\boldsymbol{\SelfStopProb} \equiv \Pr(S | \boldsymbol{\DontKnowProb})$ (Equation~\ref{eq:stoptime}).
We then define a negative ELBO loss,
\[
\ell_\text{CYB-VA} =
-\mathbb{E}_{i \sim \text{Cat}({\boldsymbol \SelfStopProb})} \log \left[ \discount_i \NextTokenProb_i \right] + D_{\text{KL}}( \boldsymbol{s}~||~{\boldsymbol{\DesiredStopProb}})  
~.
\]
Minimizing  $\ell_\text{CYB-VA}$, as an upper bound on our desired loss $\ell_{\text{CYB}\DesiredStopProb}$,  
trains the model's predictions ($\boldsymbol t$) by optimizing the expected discounted likelihood under $\boldsymbol\SelfStopProb$.
It also trains the model's stopping time distribution ($\boldsymbol \SelfStopProb$) toward the posterior, following the optimization view of Bayesian inference \citep[e.g.,][]{zellner1988}.
We can determine the equilibrium self-stop distribution by taking the gradient of the variational loss,
\[
\nabla_{\boldsymbol \SelfStopProb} ~\ell_\text{CYB-VA} = -\log\boldsymbol\discount - \log\boldsymbol t - \log\boldsymbol\DesiredStopProb + \log\boldsymbol \SelfStopProb + \boldsymbol1 ~.
\]
Introducing a Lagrange multiplier for the constraint $\sum_i \SelfStopProb_i = 1$ yields the optimality condition $\log\boldsymbol \SelfStopProb = \log\boldsymbol\discount + \log\boldsymbol t + \log\boldsymbol\DesiredStopProb - (\lambda+1)\boldsymbol1$,
resulting in $\SelfStopProb_i \propto \DesiredStopProb_i \discount_i \NextTokenProb_i$.
The model will thus learn to stop at times when both the stopping prior and its discounted accuracy are high.
This analysis also indicates that the prior, $\boldsymbol \DesiredStopProb$, and the discount factors, $\boldsymbol \discount$, are 
redundant in CYB-VA. We thus assume $\discount_i=1$ in all experiments with CYB-VA. CYB-VA thus has  hyperparameters 
$\boldsymbol \DesiredStopProb$, in addition to specifying $\MaxWorldStopStep$, the upper limit on the number of steps.

\subsection{Distributional penalty (CYB-DP)}

A final approach is to optimize for our original loss, $\mathcal{L}_\text{CYB}$ (Equation~\ref{eq:loss}), and impose a penalty based on the deviation of 
the model's desired and observed stop-time distributions:
\[
\ell_{\text{CYB-DP}} = \ell_{\text{CYB}} + \alpha~ D_{\text{KL}}(\boldsymbol{\DesiredStopProb} 
~||~\boldsymbol{s})  ~,
\]
where $\alpha$ specifies a penalty on the distribution mismatch.
Note that the direction of the KL terms is reversed for \ldp versus \lva, in order to strongly penalize stopping times that are outside the support of the prior.  (And we wish to allow the prior to have zero values.)
CYB-DP has hyperparameters $\boldsymbol \DesiredStopProb$ and $\alpha$, in addition to specifying $\MaxWorldStopStep$, the upper limit on
the number of steps. 

\subsection{Results from CYB-VA and CYB-DP simulations}

\begin{figure}[tb]
\centering
\includegraphics[width=.8\linewidth,valign=t]{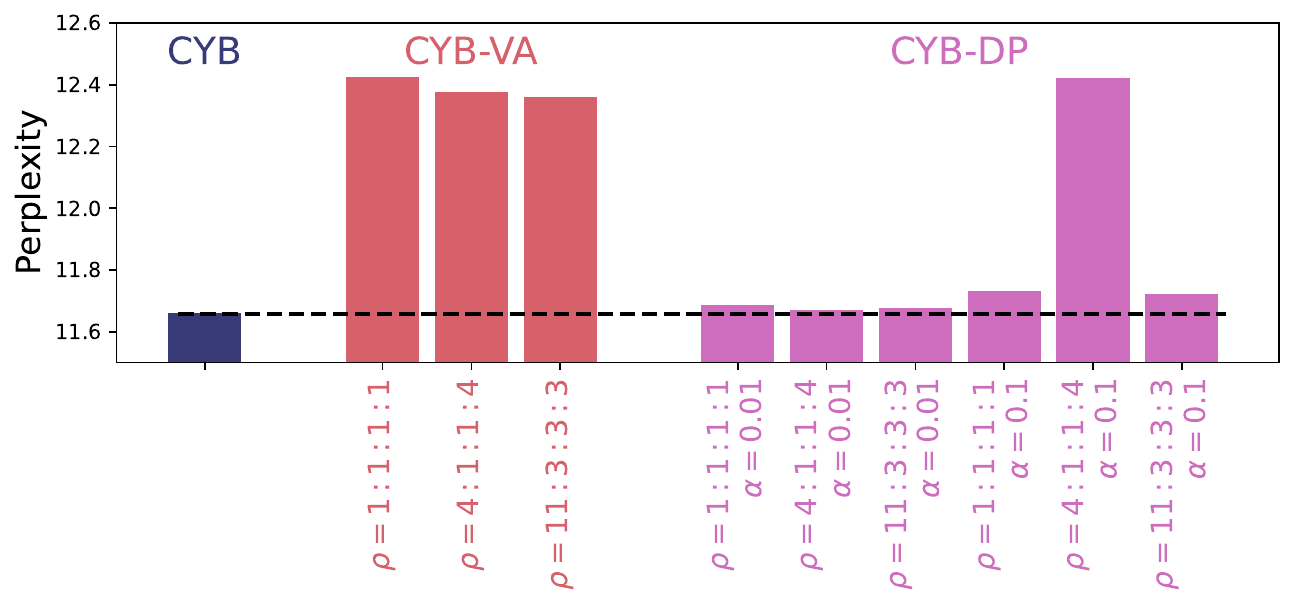}
\caption{C4 evaluation set perplexity for various hyperparameter settings of
CYB and two variants, CYB-VA and CYB-DP, on Gemma1-2B fine-tuning experiment. 
The black dashed line indicates the perplexity of CYB.
Hyperparameters $\rho$ refer to the 
distribution of stopping times and are indicated by ratios for steps 1-4 (i.e., 
the initial input token and the following 3 pause tokens). Hyperparameter
$\alpha$ is a penalty term associated with CYB-DP.
See Table \ref{table:notation} for notation.
\label{fig:hyperparameters_exploration}}
\end{figure}

\paragraph{Perplexity.}
Figure~\ref{fig:hyperparameters_exploration} shows perplexity for CYB, three hyperparameter settings for
CYB-VA and six for CYB-DP on Gemma1-2B fine-tuning experiment. 
CYB-VA performs poorly for three different target stop-time distributions, $\rho$.
We considered incorporating a scaling coefficient on the KL term, but CYB-VA's elegance stems
from the fact that such a coefficient should not be required. 
For CYB-DP, which penalizes deviation from a target stop-time distribution,
we tested two penalty coefficients, $\alpha$, and
three target distributions, $\rho$.
The smaller penalty yielded better performance, and the similar performance over
$\rho$ suggests that the penalty had little impact.  If the penalty is not shaping
learning, CYB-DP becomes equivalent to CYB with no world-induced early stopping.

\begin{figure}[b!]
    \centering
    \includegraphics[height=2in]{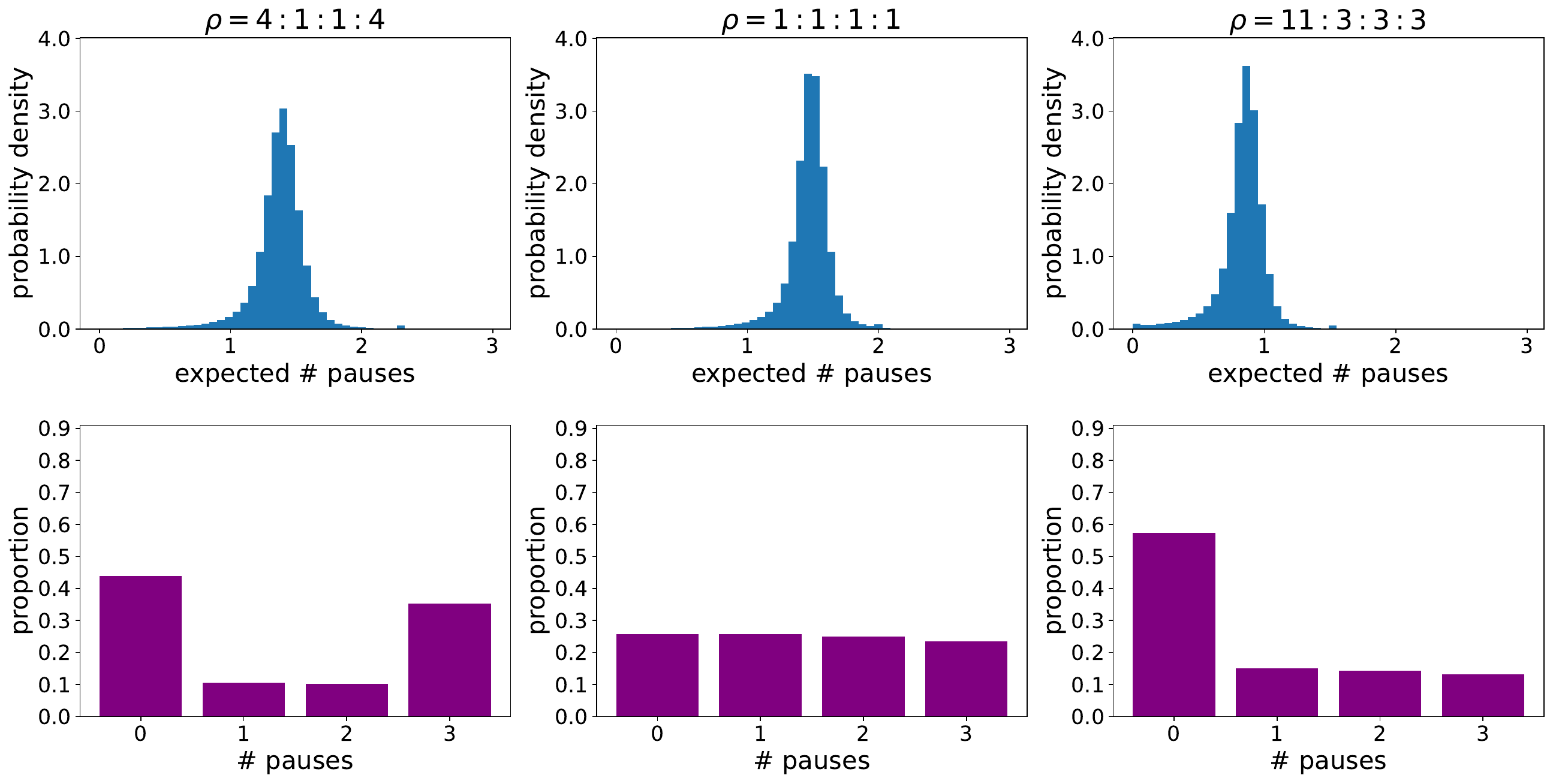}
    \caption{Distribution of stopping times for CYB-VA. Blue histograms show the distribution over evaluation tokens of the expected number of pause steps. Purple bar graphs show the distribution over the number of pause steps across tokens.}
    \label{fig:all_steps_va}
\end{figure}

\begin{figure}[tb]
    \centering
    \includegraphics[height=2in]{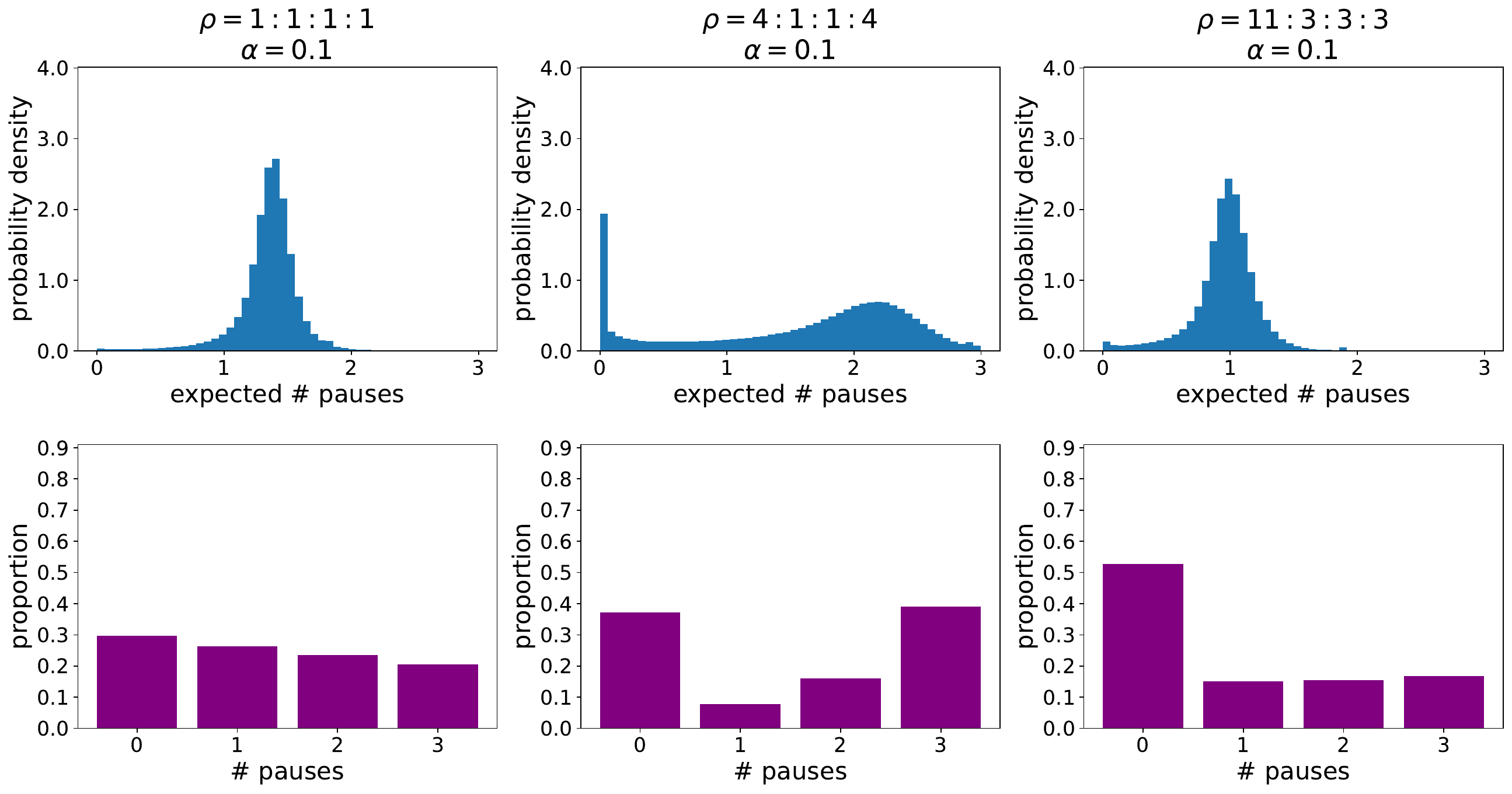}
    \includegraphics[height=2in]{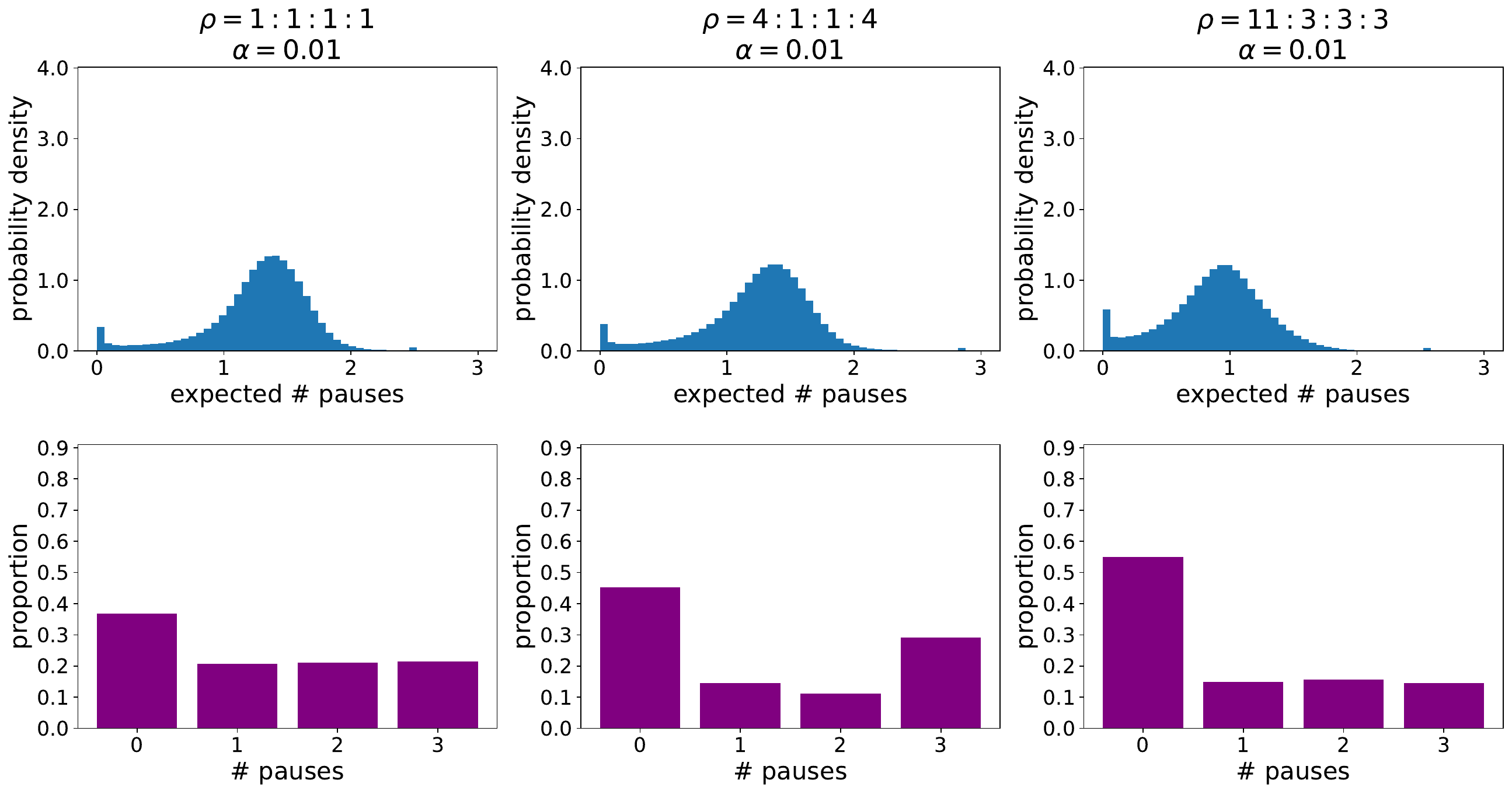}
    \caption{Distribution of stopping times for CYB-DP with $\alpha=0.1$ and $\alpha=0.01$. Blue histograms show the distribution over evaluation tokens of the expected number of pause steps. Purple bar graphs show the distribution over the number of pause steps across tokens.}
    \label{fig:all_steps_dp}
\end{figure}

\paragraph{Distributions of stopping times.} Stopping-time distributions for CYB-DP and CYB-VA are presented in 
Figures~\ref{fig:all_steps_va} and \ref{fig:all_steps_dp}. These Figures are analogous to 
Figure~\ref{fig:steps_ap} from the main paper.
CYB-VA adheres to the prior distribution and in general shows little variance indicating that it is
insensitive to the specific token and context, which is a bad sign for an adaptive-pause method. 
CYB-DP for large $\alpha=0.1$ behaves closer to CYB-VA and for smaller $\alpha=0.01$ prefers longer latencies
than CYB. However, longer latencies do not result in a lower perplexity compared to CYB.

\clearpage
\section{Additional methodology}
\label{app_sec:methods}

\subsection{Training hyperparameters and data}

For all the fine-tuning experiments, we fine-tuned either Gemma1-2B or Gemma3-4B models on a subset of $6.4$ million sequences of the C4 dataset. For Gemma1-2B model we used batch size of $256$, leading to a total of $25k$ iterations, while for Gemma3-4B model we used batch size of $64$ leading to a total of $100$k iterations. For pretraining experiments, we trained from scratch Gemma1-2B with batch size of $256$ on $56 \times 6.4$ million sequences of the C4 dataset. For the evaluation, we use 160K sequences (59.8M tokens) from the C4 validation set.


We use a ``concatenate then split'' packing algorithm and a cosine learning rate decay with a linear warmup. We select a learning
rate multiplier $\eta_0$ and obtain a maximum learning rate by dividing $\eta_0$ by 
the square root of the model embedding dimension. The minimum learning rate is given by multiplying the maximum learning rate by $10^{-2}$. Before running all the experiments, we ran a sweep over $\eta_0 \in [10^{-2}, 5\times 10^{-3}, 10^{-3},
5\times 10^{-4}, 5\times 10^{-5}, 10^{-5}, 10^{-6}]$ where we ran the fine-tuning experiment on Gemma1-2B with baseline, TBYS with 3 pauses and CYB with 3 pauses. We found that $\eta_0=10^{-4}$ gave overall the best results for all the methods and this is what we used.

All the experiments are conducted on TPUv5 hardware.





\subsection{Initializing \texorpdfstring{\dk}{don't know}}
Because the pretrained model was never rewarded for assigning probability mass to the \dk token, we wanted to ensure that the model actually had the potential to output \dk when being fine-tuned with the CYB loss. To achieve this objective, model posteriors were renormalized to reflect a high prior for \dk.  If $\boldsymbol{\psi}$ is the prior distribution over tokens  during pretraining and $\boldsymbol{\psi}'$ is the prior we wish to be reflected in the  model posteriors, we add $\log(\boldsymbol{\psi}'/\boldsymbol{\psi})$ to the model logit vector.  With uniform pretraining priors, i.e., $\psi_i= 1/|V|$ with $V$ being the vocabulary, we set  $\psi'_{\textsc{dk}}$ for the \dk and  $\psi'_i=(1-\psi'_{\textsc{dk}})/(|V|-1)$ for all  other indices $i$. In the reported experiments, we use $\psi'_{\textsc{dk}}= 0.9$.  We found the models trained better with $\psi'_{\textsc{dk}} \geq 0.9$.

\paragraph{Comparison models.} Our primary aim is to argue that CYB outperforms the memory- and compute-matched TBYS in its ability to utilize \pause tokens. TBYS is identical but is trained by the ordinary
cross-entropy loss. For TBYS, we used the same position coding scheme as in the CYB models, even 
though  the original TBYS model treated pauses like any other token for the purpose of position coding. 
(The original TBYS model also does not encode the first, second, etc. pause in sequence distinctly from 
the others.) We varied the number of pauses after each real token from one to three for both CYB and TBYS.
Evaluation of all comparison models and CYB used the same subset of 160k sequences from a validation
set of C4.

\paragraph{Baselines.} We tested a  \emph{baseline} model with no pause tokens inserted in the sequences.
The baseline model thus had the same number of real tokens per input sequence but $1/\MaxWorldStopStep$ the
overall length of the CYB and TBYS models with $\MaxWorldStopStep-1$ pauses. Thus, the baseline
is data matched but lower in memory and compute demands. We included a second baseline, which we
refer to as \emph{baseline 4$\times$seq}, which is memory and compute matched but its context window is filled
with $\MaxWorldStopStep=4$ times as many training tokens as CYB and TBYS.

\paragraph{Additional fine-tuning data regimes.} Besides the original fine-tuning regime, we ran a subset 
of methods (baselines and CYB) using $2\times$ and $3\times$ training data as well as using $2$ epochs on the original dataset.

\subsection{Where and how many pauses are permitted?}

Two distinct recipes can be used for training with the CYB losses. We use the first of these,
but future research could explore the second.
Both require that we specify the maximum number of steps allowed, $\MaxWorldStopStep$. 

\subsubsection{Recipe 1: Adaptive readout (fixed pause allocation)}
The first recipe, which we use in the experiments we report in the main paper, inserts $\MaxWorldStopStep-1$ pauses
in the input stream after each real input token during training.  The transformer steps are run in parallel with 
causal masking. The pause steps may or may not be used by the model, e.g., if at some step $i$ the model 
is certain ($\DontKnowProb_i \to 0$) or the world enforces a stop (${\WorldStopAtIProb}_i \to 1$), the outputs at subsequent pauses 
will not contribute to the loss.

During inference, when the transformer might be run autoregressively, we match the use of pauses at train time, 
i.e., always inserting $\MaxWorldStopStep-1$ pauses after each real input token. Because an input token and the pauses that 
follow can be processed in parallel, there is no computational cost to inserting the maximum number of pauses in the stream for every
token. Rather, the potential cost comes from the pauses cluttering the context window. Even when $\MaxWorldStopStep-1$
pauses are inserted at both training and inference, and even when the model is not pressured to reply quickly (e.g., 
$\gamma_i=1$ for all $i$), the model must still learn to use the \dk output to calibrate itself and determine when to read out.

\subsubsection{Recipe 2: Adaptive computation time (dynamic pause insertion)}
The second recipe involves a variable number of pause steps at both training and inference.
The CYB losses can accommodate variability in the number of pauses following a token at training. No modification to the loss 
needs to be made, but the \dk probability must be forced to zero on the last pause, or in response to an input token itself when 
there are no pauses. Rather than computing the expectation over world stopping times, we can sample a stopping time for each
token and insert that number of pause steps in the input stream. One might also envision training methods that insert
more pauses where they are likely to be most useful, e.g., where thinking might be required. (In the results section, 
we present evidence that pause requests by the model are token- and context-specific.)

During inference, the transformer might be run autoregressively, and we can again sample from the world stop-time distribution to
determine the number of pause steps, and sample from the model's \dk confidence to determine which of the allowed steps to read
out from.

\section{Calibration of model confidence and accuracy}
\label{app_sec:calibration}
Hypothesis \textbf{H2} states that
CYB should train models to self-calibrate: \dk should have high probability when a delay in responding will improve model output. 
Even without discounting ($\gamma_i = 1$) or forced stops ($\boldsymbol{\omega} = (0, \dots, 0, 1)$), minimizing the CYB loss 
implies the model should not produce a \dk response at step $i$ unless accuracy improves at step $i+1$.
The next two sections argue for the theoretical justification and empirical evidence supporting
this conjecture.

\subsection{Theoretical justification}
\label{app_sec:calibration_theo}

Under $\boldsymbol{\omega} = (0, \dots, 0, 1)$, Equation~\ref{eq:stoptime} simplifies to 
\[ 
s_i^k = (1 - d_i^k)\prod_{j=1}^{i-1} d_j^k \text{  with  } d_M^k \equiv 0.
\] 
Defining the expected discounted accuracy recursively as 
\[
V_i^k = (1 - d_i^k)\gamma_i t_i^k + d_i^k V_{i+1}^k \text{  with  } V_M^k = \gamma_M t_M^k
\text{  and  }
V_1^k = \mathbb{E}_{i \sim \Pr(S \mid \boldsymbol{d}^k, \boldsymbol{\omega})}[\gamma_i t_i^k],
\]
the partial derivative of $\ell_{\text{CYB}} = -\log V_1^k$ with respect to the \dk output is
$$\frac{\partial \ell_{\text{CYB}}}{\partial d_i^k} = -\frac{\prod_{j=1}^{i-1} d_j^k}{V_1^k}\left(V_{i+1}^k - \gamma_i t_i^k\right) \quad$$
Critically, the sign of the derivative depends not on accuracy alone, but on whether the marginal future gain ($V_{i+1}^k$) 
is larger than the current discounted accuracy ($\gamma_i t_i^k$). Increasing the \dk probability will reduce the CYB loss 
when abstention will lead to a better outcome on expectation. Note that $\gamma_i<1$ will encourage abstention only when 
absolute accuracy is increasing. Without discounting, $d_i^k$ will be driven low only for tokens where 
subsequent pause steps actively degrade expected target accuracy.

\subsection{Empirical evidence}
\label{app_sec:calibration_emp}
To be successful, the model must implicitly predict its own  accuracy at subsequent steps.  That is, the probability of responding \dk at step $i$, $d_i$, should 
be large when the target probability is higher at the next step, i.e., $t_{i+1} > t_i$. 

\begin{figure}[htb]
    \centering
    \includegraphics[width=\linewidth]{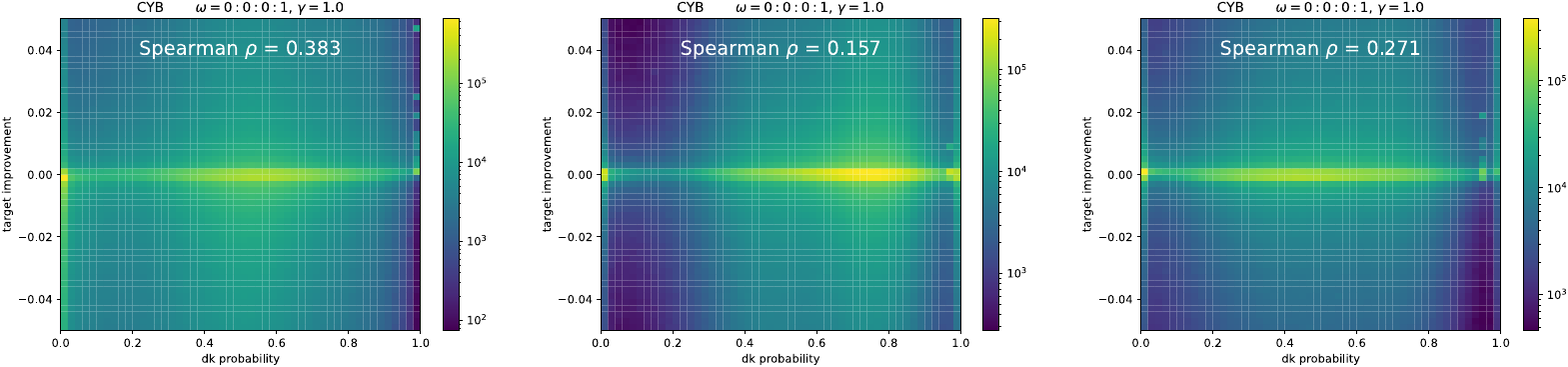}
    \caption{Density plots indicating that \dk probability is positively correlated with improvement in accuracy, for steps 1, 2, and 3.} \label{fig:dk_vs_improvement}
\end{figure}

Figure~\ref{fig:dk_vs_improvement} shows joint density maps of \dk probability (along abscissa) and the improvement in accuracy
obtained by a one-step delay in responding (along ordinate). The three density maps correspond to steps $i \in \{1,2,3\}$.
A positive correlation indicates that the model can predict when it will benefit by the delay. Over the evaluation  set, 
the Spearman correlation coefficient between  $d_i$ and $t_{i+1}-t_i$ is 0.383, 0.157, and 0.271 for steps 1-3, respectively,
all reliably nonzero ($p < 0.001$).\footnote{The statistical reliability of the non-zero 
correlation coefficient is partly due to the large sample size used to compute the correlation.} The correlations are only slightly stronger with the discounting variant of CYB ($\gamma = 
0.99^{\{0,1,2,3\}}$)---0.381, 0.208, and 0.309---indicating that penalizing delayed responses does not appear to make the model
much better at determining when to pause.

\clearpage
\section{Additional results}
\label{sec_app:additional_results}

In this section, we present additional results.

\subsection{Tabular representations of the results from the main paper}
\label{sec_app:tabular_representation}

In this section, we present the results from Figure~\ref{fig:finetuning_results} and Figure~\ref{fig:pretraining_performance} in a tabular format for the easy introspection. We report C4 eval set perplexity as well as the relative perplexity improvement (in \%) compared to the baseline perplexity. Relative improvement is computed as the percentage decrease in perplexity relative to the TBYS with 3 pause tokens, i.e.
\begin{equation}
    \text{Relative Improvement} = (\text{PPL}_{\text{TBYS,p=3}} - \text{PPL}_{\text{method}}) / \text{PPL}_{\text{TBYS,p=3}} \times 100\%
    \label{eq:rel_improvement}
\end{equation}

The results for fine-tuning of Gemma3-4B model are given in Table~\ref{tab:gemma4b_3}. We see that even CYB with 1 pause outperforms TBYS with 3 pauses. Using 3 pauses with CYB improves performance even further leading to \textbf{+2.2\%} relative improvement in perplexity. The results for fine-tuning of Gemma1-2B model are given in Table~\ref{tab:gemma1_2b_finetuning}. Overall, the behavior is very similar to Gemma3-4B experiment. Finally, the results of pretraining of Gemma1-2B model are given in Table~\ref{tab:gemma1_2b_pretraining}. What is remarkable is that using CYB with the pretrained model leads to even higher relative improvement in perplexity, notably \textbf{+3.57\%} compared to the fine-tuned model which enjoys \textbf{+2.72\%} relative improvement.


\begin{table}[hbt]
\centering
\begin{tabular}{lrr}
\toprule
\textbf{Method} & \textbf{Perplexity} & \textbf{Rel. Improvement (\%)} \\
\midrule
Pretrained Gemma3-4B & 11.765 & -13.57\% \\
\hline
Baseline & 10.670 & -3.00\% \\
Baseline 4$\times$seq & 10.694 & -3.24\% \\
\hline
TBYS 1 pause & 10.512 & -1.48\% \\
TBYS 2 pauses & 10.422 & -0.61\% \\
TBYS 3 pauses & 10.359 & --- \\
\hline
CYB 1 pause & 10.351 & +0.08\% \\
CYB 2 pauses & 10.210 & +1.44\% \\
CYB 3 pauses & \textbf{10.131} & \textbf{+2.20\%} \\
\bottomrule
\end{tabular}
\caption{\textbf{Results for fine-tuned (on C4) Gemma3-4B}. Perplexity on eval set of C4. For relative improvement, we use the formula~\eqref{eq:rel_improvement}. In \textbf{bold} we highlight the best result -- the lowest perplexity and the highest relative improvement.}
\label{tab:gemma4b_3}
\end{table}


\begin{table}[htb]
\centering
\small
\begin{tabular}{lrr}
\toprule
\textbf{Method} & \textbf{Perplexity} & \textbf{Rel. Improvement (\%)} \\
\midrule
Pretrained Gemma1-2B & 13.018 & -8.64\% \\
\hline
Baseline & 12.173 & -1.58\% \\
Baseline 4$\times$seq & 11.953 & +0.25\% \\
\hline
TBYS 1 pause & 12.098 & -0.96\% \\
TBYS 2 pauses & 12.024 & -0.34\% \\
TBYS 3 pauses & 11.983 & --- \\
\hline
CYB 1 pause & 11.865 & +0.99\% \\
CYB 2 pauses & 11.758 & +1.88\% \\
CYB 3 pauses & \textbf{11.657} & \textbf{+2.72\%} \\
\bottomrule
\end{tabular}
\caption{\textbf{Results for fine-tuned (on C4) Gemma1-2B}. Perplexity on eval set of C4. For relative improvement, we use the formula~\eqref{eq:rel_improvement}. In \textbf{bold} we highlight the best result -- the lowest perplexity and the highest relative improvement.}
\label{tab:gemma1_2b_finetuning}
\end{table}

\begin{table}[hbt]
\centering
\begin{tabular}{lrr}
\toprule
\textbf{Method} & \textbf{Perplexity} & \textbf{Rel. Improvement (\%)} \\
\midrule
Pretrained Gemma1-2B & 13.018 & -4.14\% \\
\hline
Baseline & 12.893 & -3.14\% \\
\hline
TBYS 3 pauses & 12.500 & --- \\
\hline
CYB 3 pauses & \textbf{12.054} & \textbf{+3.57\%} \\
\bottomrule
\end{tabular}
\caption{\textbf{Results for trained from scratch (on C4) Gemma1-2B}. Perplexity on eval set of C4. For relative improvement, we use the formula~\eqref{eq:rel_improvement}. In \textbf{bold} we highlight the best result -- the lowest perplexity and the highest relative improvement.}
\label{tab:gemma1_2b_pretraining}
\end{table}



\begin{figure}[hbt]
    \centering
    \includegraphics[width=5.5in]{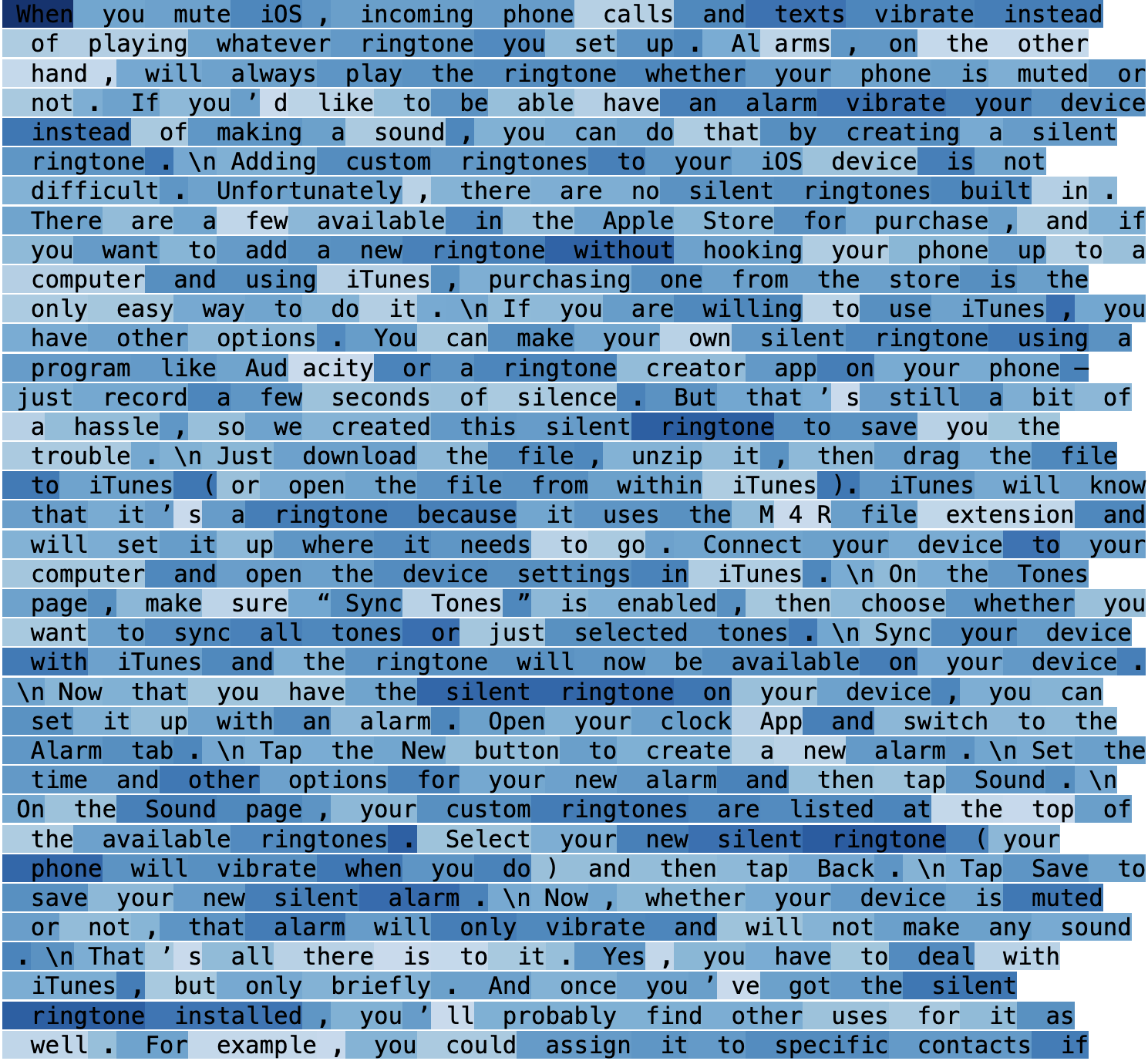}
    \includegraphics[width=2.7in]{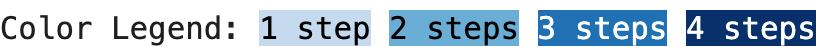}
    \caption{Model trained with CYB loss produces variable
    pause durations on tokens.} \label{fig:test_example2}
\end{figure}
\begin{figure}[hbt]
    \centering
    \includegraphics[width=5.5in]{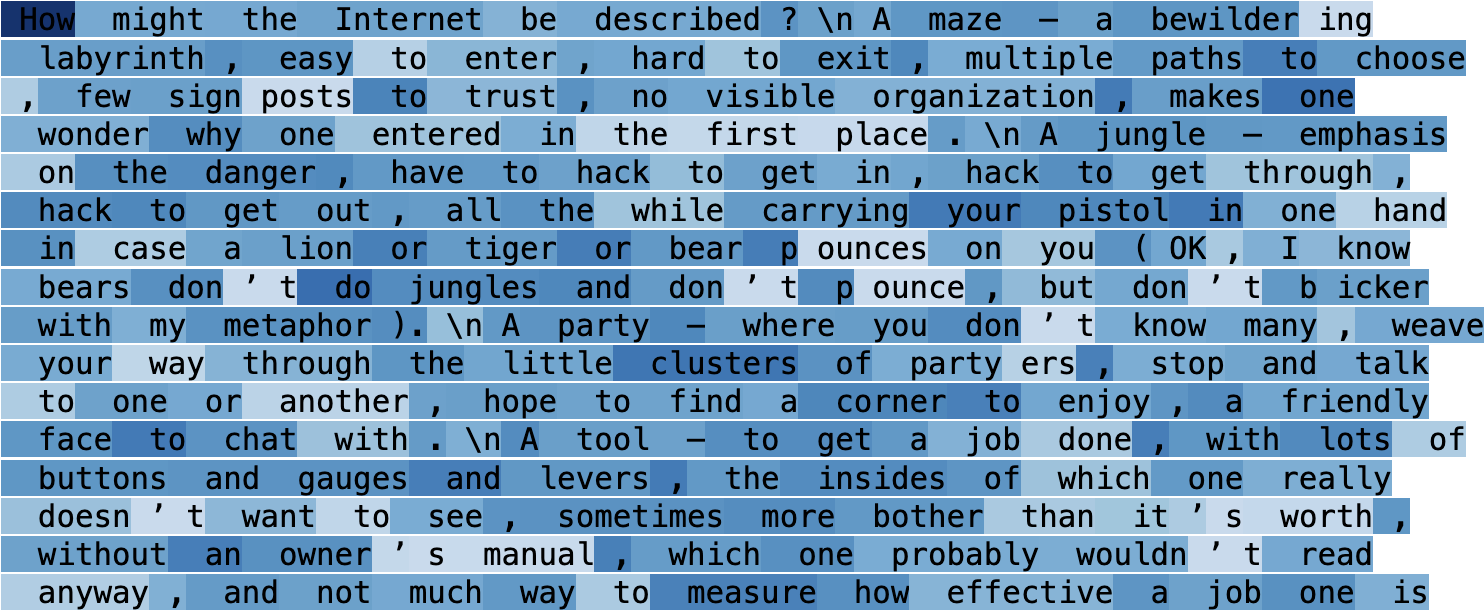}
    \includegraphics[width=2.7in]{fig/text_legend.png}
    \caption{Model trained with CYB loss produces variable
    pause durations on tokens.} \label{fig:test_example3}
\end{figure}

\newpage
\section{Derivation of output probability}
\label{app_sec:derivation_output_prob}

Based on Figure~\ref{fig:mdp}, the probability of outputting a result at step $i$ is given by
\begin{equation}
    \Pr(\SelfStopStep=i | \boldsymbol{\DontKnowProb},\boldsymbol{\WorldStopProb}) = \left(\prod_{j=1}^{i-1} \DontKnowProb_j \right) \left( \prod_{j=1}^{i-1} (1- \WorldStopAtIProb_j) \right) \left( \DontKnowProb_i \WorldStopAtIProb_i + (1-\DontKnowProb_i) \right)
    \label{eq_app:prob_stop}
\end{equation}

Using the definition of the survival probability in Figure~\ref{fig:mdp},
\begin{equation}
\tilde{\omega}_i = {\WorldStopProb_i}/{\textstyle \sum_{j=i}^{\MaxWorldStopStep} \WorldStopProb_j}
\end{equation}
we notice that
\begin{align*}
    \DontKnowProb_i \WorldStopAtIProb_i + (1-\DontKnowProb_i) &= \frac{1}{\sum_{k=i}^{\MaxWorldStopStep} \WorldStopProb_k} \left( \DontKnowProb_i \WorldStopProb_i + (1- \DontKnowProb_i) \sum_{k=i}^{\MaxWorldStopStep} \WorldStopProb_k \right) \\
    &= \frac{1}{\sum_{k=i}^{\MaxWorldStopStep} \WorldStopProb_k} \left( \DontKnowProb_i \WorldStopProb_i + \WorldStopProb_i - \DontKnowProb_i \WorldStopProb_i + (1- \DontKnowProb_i) \sum_{k=i+1}^{\MaxWorldStopStep} \WorldStopProb_k \right) \\
    &= \frac{1}{\sum_{k=i}^{\MaxWorldStopStep} \WorldStopProb_k} \left( \WorldStopProb_i + (1- \DontKnowProb_i) \sum_{k=i+1}^{\MaxWorldStopStep} \WorldStopProb_k \right) 
\end{align*}
Moreover, we also notice that
\begin{align}
    \prod_{j=1}^{i-1} (1- \WorldStopAtIProb_j) &= \prod_{j=1}^{i-1} \frac{\sum_{k=j+1}^{\MaxWorldStopStep} \WorldStopProb_k}{\sum_{k=j}^{\MaxWorldStopStep} \WorldStopProb_k} \\
    &= \frac{\prod_{j=1}^{i-1} \sum_{k=j+1}^{\MaxWorldStopStep} \WorldStopProb_k}{\prod_{j=1}^{i-1} \sum_{k=j}^{\MaxWorldStopStep} \WorldStopProb_k} \\
    &= \left( \sum_{k=i}^{\MaxWorldStopStep} \WorldStopProb_k \right) \left(\frac{1}{\sum_{k=1}^{\MaxWorldStopStep} \WorldStopProb_k} \right)  \left( \frac{\prod_{j=2}^{i-1} \sum_{k=j}^{\MaxWorldStopStep} \WorldStopProb_k}{\prod_{j=2}^{i-1} \sum_{k=j}^{\MaxWorldStopStep} \WorldStopProb_k} \right) \\
    &= \sum_{k=i}^{\MaxWorldStopStep} \WorldStopProb_k,
\end{align}
where we used the fact that $\sum_{k=1}^{\MaxWorldStopStep} \WorldStopProb_k = 1$. Now, we plug these two into~\eqref{eq_app:prob_stop} and we get
\begin{equation}
    \Pr(\SelfStopStep=i | \boldsymbol{\DontKnowProb},\boldsymbol{\WorldStopProb}) = \left(\prod_{j=1}^{i-1} \DontKnowProb_j \right) \left( \WorldStopProb_i + (1-\DontKnowProb_i) \sum_{k=i+1}^{\MaxWorldStopStep} \WorldStopProb_k\right)
    \label{eq_app:prob_stop_proved}
\end{equation}

\end{document}